\documentclass[10pt,twocolumn,letterpaper]{article}
\PassOptionsToPackage{table}{xcolor}   %

\usepackage[pagenumbers]{cvpr} %

\usepackage{multirow}
\usepackage{makecell}
\usepackage{pifont}       %
\usepackage{xspace}
\usepackage{amsthm}
\usepackage{float}
\usepackage{titletoc}   %
\titlecontents{section}[1.6em]{\addvspace{6pt}\bfseries}{\contentslabel{1.6em}}{}{\hfill\contentspage}
\titlecontents{subsection}[3.9em]{\addvspace{1pt}}{\contentslabel{2.3em}}{}{\titlerule*[0.6em]{.}\contentspage}

\newtheorem{proposition}{Proposition}

\usepackage{algpseudocode}\algrenewcommand\algorithmiccomment[1]{\hfill{\color{gray}$\triangleright$ #1}}
\floatstyle{ruled}\newfloat{algorithm}{t}{loa}\floatname{algorithm}{Algorithm}

\definecolor{ourgreen}{RGB}{225,245,225}

\newif\ifshowrev\showrevfalse
\definecolor{revblue}{RGB}{0,70,200}
\makeatletter\long\def\rev#1{\ifshowrev{\color{revblue}#1}\else#1\fi}\makeatother
\newcommand{\revcolor}{\ifshowrev\color{revblue}\fi}
\definecolor{revred}{RGB}{200,0,0}  %


\definecolor{clred}{RGB}{214,30,80}


\definecolor{pendorange}{RGB}{205,90,0}


\newcommand{\method}{\rev{DecoFlat}\xspace}  %
\newcommand{\VAS}{orthogonal perturbation\xspace}   %
\newcommand{\CAU}{conflict-aware correction\xspace}   %
\newcommand{\SHR}{color rescheduling\xspace}   %

\definecolor{okgreen}{RGB}{0,140,60}
\definecolor{nored}{RGB}{200,30,30}
\newcommand{\cmark}{\textcolor{okgreen}{\ding{51}}}
\newcommand{\xmark}{\textcolor{nored}{\ding{55}}}
\newcommand{\sd}[1]{{\scriptsize$\pm$#1}}

\newcommand{\gv}{\mathbf{g}}
\newcommand{\gbar}{\bar{\mathbf{g}}}
\newcommand{\gperp}{\mathbf{g}_{\perp}}
\newcommand{\gpar}{\mathbf{g}_{\parallel}}
\newcommand{\tgv}{\tilde{\mathbf{g}}}
\newcommand{\dv}{\mathbf{d}}
\newcommand{\uv}{\mathbf{h}}  %
\newcommand{\params}{\boldsymbol{\theta}}
\newcommand{\pert}{\boldsymbol{\delta}}

\usepackage{tikz}
\definecolor{cnumA}{RGB}{224,161,6}\definecolor{cnumB}{RGB}{46,139,87}\definecolor{cnumC}{RGB}{214,46,46}
\DeclareRobustCommand{\cnum}[2]{\tikz[baseline=(c.base)]\node[circle,fill=#2,draw=white,line width=0.3pt,text=white,inner sep=0.7pt,font=\fontsize{6}{6}\selectfont\bfseries\sffamily](c){#1};}

\definecolor{cvprblue}{rgb}{0.21,0.49,0.74}
\usepackage[pagebackref,breaklinks,colorlinks,allcolors=cvprblue]{hyperref}
\hypersetup{pdftitle={Not All Directions Are Flat: Decomposed-Gradient Flatness for Sparse-View 3D Gaussian Splatting},pdfauthor={Ungbae Yoon, Kwanyoung Kim}}

\def\confName{CVPR}
\def\confYear{2027}

\title{\rev{Not All Directions Are Flat: Decomposed-Gradient Flatness\\ for Sparse-View 3D Gaussian Splatting}}

\author{Ungbae Yoon, \quad Kwanyoung Kim$^{\dagger}$\\
Department of AI, Gwangju Institute of Science and Technology (GIST)
}

\begin{document}
\twocolumn[{%
  \renewcommand\twocolumn[1][]{#1}%
  \maketitle
  \begin{center}
    \vspace{-1.6em}
    \includegraphics[width=1.0\textwidth]{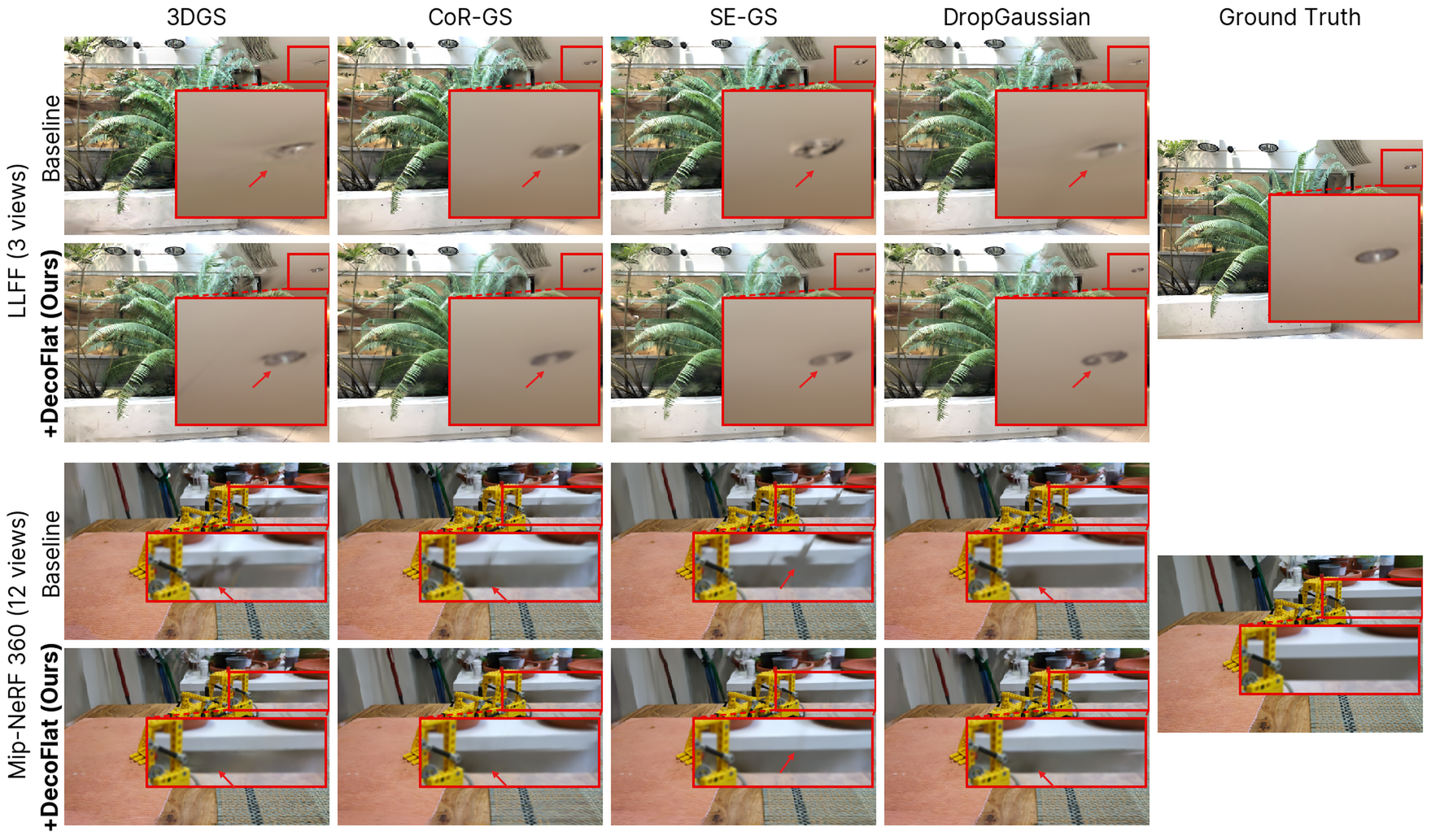}\vspace{-0.6em}
    \captionof{figure}{\rev{\textbf{Novel views without and with \method (Ours)} (LLFF, 3 views; Mip-NeRF\,360, 12 views). As a plug-in optimizer for vanilla 3DGS~\cite{kerbl2023gaussian}, CoR-GS~\cite{zhang2024corgs}, SE-GS~\cite{segs2025}, and DropGaussian~\cite{park2025dropgaussian}, \method improves the novel views of every baseline and recovers fine structures that the baseline distorts, without changing its representation, renderer, or supervision.}}
    \label{fig:teaser}
    \vspace{-0.4em}
  \end{center}
}]
{\renewcommand{\thefootnote}{\fnsymbol{footnote}}\footnotetext[2]{Corresponding author: \texttt{k0.kim@gist.ac.kr}}}
\begin{abstract}
\rev{Many applications of novel view synthesis, such as augmented reality and robotics, can afford only a few input views, since dense capture is costly. With such sparse views, however, 3D Gaussian Splatting (3DGS) overfits the training views and degrades at unseen viewpoints. Among the approaches to this problem, flat-minima optimization treats the Gaussians as trainable parameters and steers training toward flat minima, which generalize better, by perturbing the Gaussians with random noise or along the training gradient. However, these perturbations ignore their direction relative to the other training views. Interestingly, we find that the direction of the perturbation determines whether it improves generalization. Specifically, we decompose the gradient of each training view into a component parallel to the \rev{average} gradient of all views, on which the views agree, and an orthogonal component, on which they disagree. Perturbing along the orthogonal component improves generalization, whereas perturbing along the parallel component does not. \rev{We further find that the flatness correction often opposes the reconstruction gradient, shrinking or even reversing the reconstruction step.} Based on \rev{these findings}, we propose \textbf{\method}, which perturbs each Gaussian \rev{along the orthogonal component together with random noise}. Furthermore, we introduce a conflict-aware \rev{correction}, which \rev{removes} the part of the flatness correction that \rev{opposes} the reconstruction gradient, and a rescheduling of the spherical-harmonics color, which suppresses the overfitting of view-dependent appearance. Extensive experiments demonstrate that \method, as a plug-in optimizer, consistently improves vanilla 3DGS and a wide range of sparse-view 3DGS methods.}
\end{abstract}

\section{\rev{Introduction}}
\label{sec:intro}
\begin{table}[t]
\centering\revcolor
\caption{\textbf{Comparison with related methods.} Only \method chooses the perturbation direction from the gradients of all views.}
\label{tab:position}
\footnotesize
\setlength{\tabcolsep}{2pt}
\begin{tabular}{lccccc}
\toprule
Method & \makecell{For\\3DGS} & \makecell{Flat\\minima} & \makecell{Gradient\\decomp.} & \makecell{Other\\views} & \makecell{Conflict-\\aware} \\
\cmidrule(r){1-1}\cmidrule(lr){2-6}
SAM~\cite{foret2021sam} & \xmark & \cmark & \xmark & \xmark & \xmark \\
GSAM~\cite{zhuang2022gsam} & \xmark & \cmark & \cmark & \xmark & \xmark \\
LookSAM~\cite{liu2022looksam} & \xmark & \cmark & \cmark & \xmark & \xmark \\
3DGS-MCMC~\cite{kheradmand2024mcmc} & \cmark & \xmark & \xmark & \xmark & \xmark \\
GDAGS~\cite{zhou2026gdags} & \cmark & \xmark & \xmark & \xmark & \xmark \\
SAP~\cite{seo2026sap} & \cmark & \cmark & \xmark & \xmark & \xmark \\
FASR~\cite{yun2026flat} & \cmark & \cmark & \xmark & \xmark & \xmark \\
\cmidrule(r){1-1}\cmidrule(lr){2-6}
\rowcolor{ourgreen}
\textbf{\method (Ours)} & \cmark & \cmark & \cmark & \cmark & \cmark \\
\bottomrule
\end{tabular}
\end{table}

\rev{Photorealistic novel view synthesis from images has advanced rapidly.
Neural radiance fields (NeRF)~\cite{mildenhall2020nerf} first achieved it from dense captures, and 3D Gaussian Splatting (3DGS)~\cite{kerbl2023gaussian} has since become the mainstream representation because it renders in real time at comparable quality, representing a scene as a set of anisotropic Gaussians optimized by differentiable rasterization.
Many applications, such as augmented reality and robotics, however, can afford only a few input views.
With such sparse views, NeRF recovers wrong geometry~\cite{niemeyer2022regnerf} and 3DGS overfits the training views~\cite{zhu2024fsgs}.
\rev{In 3DGS, a Gaussian observed from only a few directions can fit the training images at a position that no other view supports.}
Sparse views thus produce floaters, spurious Gaussians in empty space, and distorted geometry at novel viewpoints.}

\rev{Existing approaches to sparse-view 3DGS fall into several groups.
Some supply information that the images lack, such as monocular depth~\cite{li2024dngaussian,zhu2024fsgs} or pseudo views~\cite{zhu2024fsgs}.
Others regularize training without external priors, by co-regularizing multiple models~\cite{zhang2024corgs,segs2025} or by randomly dropping Gaussians~\cite{park2025dropgaussian,dropoutgs2025}.
Since 3DGS is optimized as a set of trainable parameters, overfitting to sparse views can also be viewed as a generalization problem of the optimization.
Recent methods therefore address it at the level of the optimizer by seeking flat minima of the training loss~\cite{yun2026flat,seo2026sap}.}

\rev{Specifically, flat minima, where the loss changes little under small perturbations of the parameters, have long been associated with generalization and robustness~\cite{hochreiter1997flat,keskar2017large}, properties that matter most when the data constrain the parameters only weakly, as sparse views do.}
\rev{Two families of methods seek them.
Random weight perturbation injects noise into the parameters during training~\cite{li2024rwp}, whereas sharpness-aware minimization (SAM)~\cite{foret2021sam} perturbs the parameters along the gradient and descends from the perturbed point.
Both have counterparts in 3DGS.
Noise injected into the Gaussian parameters appears in 3DGS-MCMC~\cite{kheradmand2024mcmc} for exploration and in SAP~\cite{seo2026sap} for flatness, while FASR~\cite{yun2026flat} adapts SAM to 3DGS by setting its radius and weight from the local image frequency.
These methods thus decide how far to perturb and how much to correct, but not in which direction.
In particular, they choose the perturbation without regard to the other training views, which under sparse supervision are the only evidence of what generalizes.
\Cref{tab:position} summarizes this comparison.}

\begin{figure}[t]
  \centering
  \includegraphics[width=\linewidth]{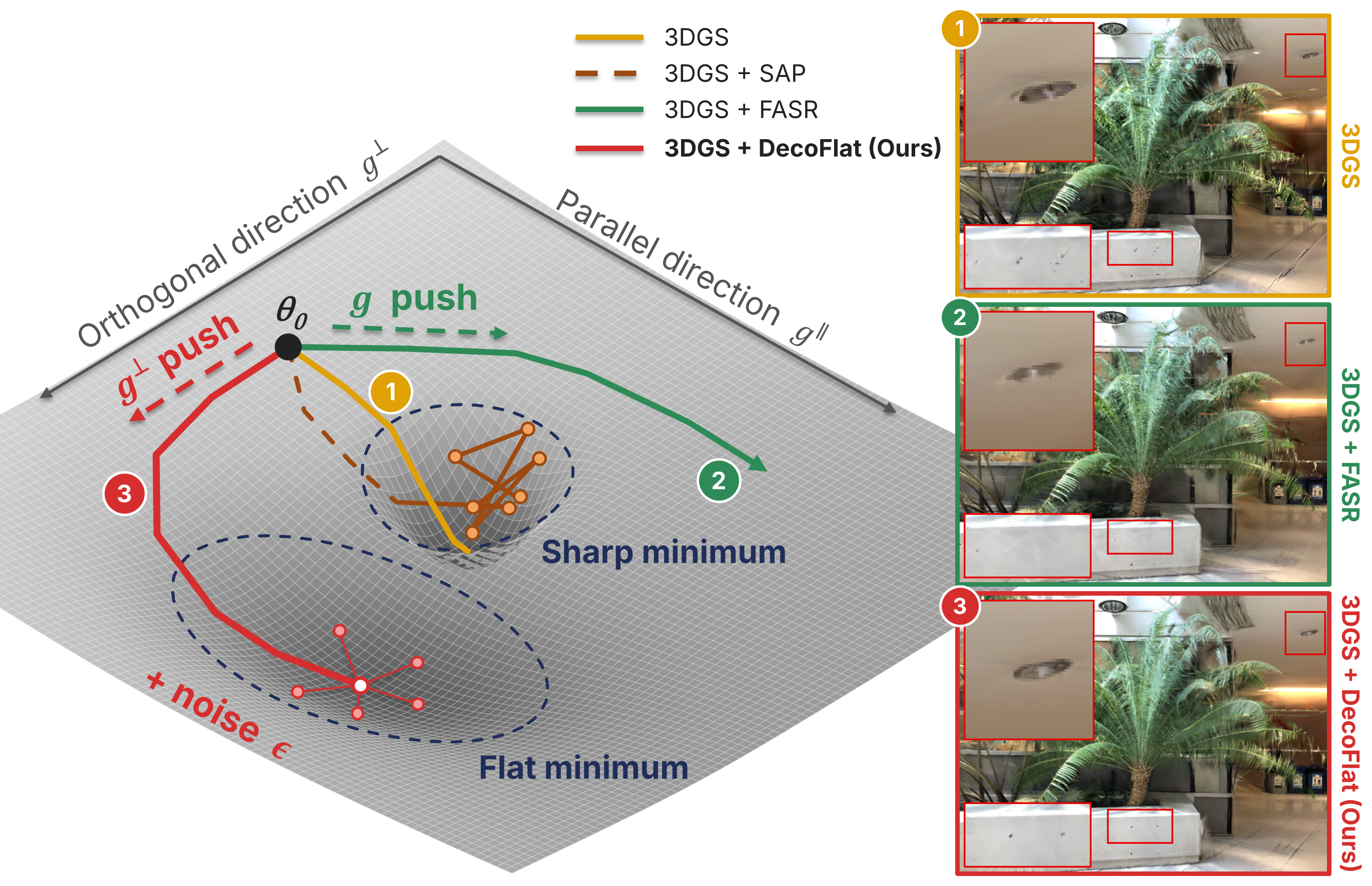}
  \caption{\rev{\rev{\textbf{Conceptual illustration of the perturbation direction.} The training loss is flat along the direction on which the views agree ($\gpar$) and sharp along the one on which they disagree ($\gperp$).} \rev{3DGS \rev{\cnum{1}{cnumA}} falls into the sharp minimum, and SAP stays inside it; FASR \rev{\cnum{2}{cnumB}} pushes the Gaussian along $\gv$; \method{} \rev{\cnum{3}{cnumC}} pushes it along $\gperp$ and reaches the flat minimum.} Right: novel views.}}
  \label{fig:toy}
\end{figure}

\rev{Interestingly, we find that the direction of the perturbation determines whether flat-minima optimization improves generalization under sparse supervision.
For a Gaussian observed by several training views, we decompose the gradient of each view into a component parallel to the \rev{average} gradient of all views, on which the views agree, and an orthogonal component, on which they disagree.
Perturbing along the orthogonal component improves generalization, whereas perturbing along the parallel component, \rev{the direction that all views share, degrades the reconstruction} as the radius grows (\cref{tab:direction}).
\Cref{fig:toy} illustrates this \rev{conceptually}: plain descent \rev{and random noise remain in the sharp minimum, and a perturbation along the raw gradient ignores the other views}, whereas a perturbation along the orthogonal component, followed by noise, reaches the \rev{flat minimum}.}

\rev{We further find that the direction matters for the update as well.
Existing flat-minima optimizers update each Gaussian along the clean gradient plus a flatness correction, the change of the gradient under the perturbation~\cite{foret2021sam,yun2026flat}.
When this correction opposes the clean gradient, a case we call a conflict, it shrinks the reconstruction step or even reverses it \rev{in a large share of the updates} (\cref{tab:e1}).}
\begin{figure*}[t]
  \centering
  \IfFileExists{figs/overview_v3.png}{\includegraphics[width=0.92\linewidth]{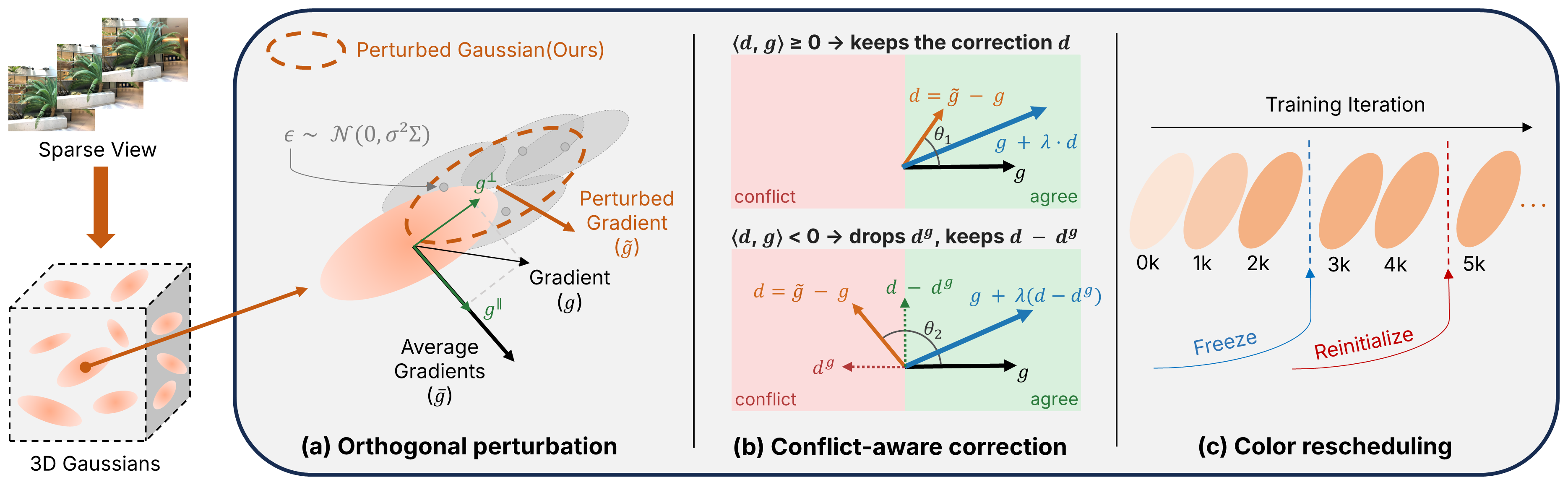}}{\fbox{\parbox[c][5cm][c]{0.9\linewidth}{\centering }}}
  \caption{\rev{\textbf{Overview of \method.} (a)~The orthogonal perturbation moves each Gaussian along the component of its gradient orthogonal to the \rev{average} gradient of all views. (b)~The conflict-aware correction drops the part of the correction $\dv$ that opposes the gradient $\gv$. \rev{(c)~Color rescheduling freezes the SH color, re-initializes it, and refits it on the settled geometry.}}}
  \label{fig:overview}
\end{figure*}

\rev{Building on these findings, we propose \method, a plug-in optimizer that revises flat-minima optimization for sparse-view 3DGS in three respects (\cref{fig:overview}).
\rev{First, since flatness helps only where the views disagree, an orthogonal perturbation moves each Gaussian only along the orthogonal component.
Second, since a conflict shrinks or reverses the reconstruction step, a conflict-aware correction removes the opposing part of the correction in that case only.}
Third, since flatness of the geometry does not prevent the view-dependent color from overfitting, \rev{a color rescheduling freezes and re-initializes the spherical-harmonics (SH) coefficients of the color} early in training.
Unlike the periodic reinitialization of SAP, which also resets the scale and rotation, this rescheduling acts only on the SH color.
\rev{None of the three changes the representation, the renderer, or the supervision.}}

\rev{As a plug-in optimizer, \method consistently improves vanilla 3DGS and a wide range of sparse-view 3DGS methods, at a training cost comparable to that of \rev{existing flat-minima optimizers}, and removes view-specific artifacts that the baselines render (\cref{fig:teaser}).
Our contributions are summarized as follows.}
\begin{itemize}[leftmargin=*,itemsep=1pt,topsep=2pt]
  \item \rev{To address the missing choice of direction in flat-minima optimization for sparse-view 3DGS, we propose \method, which decomposes the gradient of each view against the \rev{average} gradient of all views and uses this decomposition for both the perturbation and the update.}
  \item \rev{\rev{We find that the flatness correction often opposes the clean gradient, which shrinks and can even reverse the reconstruction step,} and introduce a conflict-aware correction that prevents this, together with a color rescheduling that suppresses view-dependent color overfitting.}
  \item \rev{Through extensive experiments, we show that \method consistently improves vanilla 3DGS and a wide range of sparse-view 3DGS methods as a plug-in optimizer.}
\end{itemize}

\section{\rev{Preliminaries}}
\label{sec:related}

\noindent\rev{\textbf{3D Gaussian Splatting.}}
\rev{3DGS~\cite{kerbl2023gaussian} represents a scene as a set of anisotropic Gaussians rendered by differentiable rasterization.
Gaussian $i$ has parameters $\params_i=\{\boldsymbol{\mu}_i,\mathbf{q}_i,\mathbf{s}_i,o_i,\mathbf{c}_i\}$, namely a mean $\boldsymbol{\mu}_i\in\mathbb{R}^3$, a rotation $\mathbf{q}_i$, a scale $\mathbf{s}_i$, an opacity $o_i$, and the spherical-harmonics (SH) coefficients $\mathbf{c}_i$ of its color, and is defined by}
\begin{equation}
  \rev{G_i(\mathbf{x})=\exp\!\Bigl(-\tfrac{1}{2}(\mathbf{x}-\boldsymbol{\mu}_i)^{\top}\boldsymbol{\Sigma}_i^{-1}(\mathbf{x}-\boldsymbol{\mu}_i)\Bigr),}
  \label{eq:gaussian}
\end{equation}
\rev{with covariance $\boldsymbol{\Sigma}_i=\mathbf{R}_i\mathbf{S}_i\mathbf{S}_i^{\top}\mathbf{R}_i^{\top}$, where $\mathbf{R}_i$ is the rotation matrix of $\mathbf{q}_i$ and $\mathbf{S}_i=\mathrm{diag}(\mathbf{s}_i)$.
The color of pixel $\mathbf{p}$ blends the depth-sorted set $\mathcal{N}(\mathbf{p})$ of Gaussians that cover it,}
\begin{equation}
  \rev{C(\mathbf{p})=\sum_{i\in\mathcal{N}(\mathbf{p})}\mathbf{c}_i(\mathbf{r})\,\alpha_i\prod_{j\in\mathcal{N}(\mathbf{p}),\,j<i}(1-\alpha_j),}
  \label{eq:render}
\end{equation}
\rev{where $\mathbf{c}_i(\mathbf{r})$ is the color given by $\mathbf{c}_i$ in the viewing direction $\mathbf{r}$, $\alpha_i=o_i\,G'_i(\mathbf{p})$, and $G'_i$ is the 2D projection of $G_i$.
The parameters $\params=\{\params_i\}_{i=1}^N$ are optimized by}
\begin{equation}
  \rev{\min_{\params}\ \mathcal{L}_v(\params),}
  \label{eq:update3dgs}
\end{equation}
\rev{where $\mathcal{L}_v=(1-\lambda_{\mathrm{D}})\mathcal{L}_1+\lambda_{\mathrm{D}}\mathcal{L}_{\text{D-SSIM}}$ is the photometric loss of the training view $v\in\mathcal{V}$ sampled at each iteration.}

\noindent\rev{\textbf{Sparse-view 3DGS.}}
\rev{With only a few input views, this optimization overfits the training images.
Existing methods either add priors such as monocular depth and pseudo views~\cite{li2024dngaussian,zhu2024fsgs} or regularize training by co-training multiple models~\cite{zhang2024corgs,segs2025} or by randomly dropping Gaussians~\cite{park2025dropgaussian,dropoutgs2025,dropansh2026}.
\method changes only the gradient in \cref{eq:update3dgs} and is complementary to both groups (\cref{sec:exp_plugin}).}

\noindent\rev{\textbf{Flat-minima optimization for 3DGS.}}
\rev{Flat minima of the training loss generalize better than sharp ones~\cite{hochreiter1997flat,keskar2017large}, and seeking them has been extended to various domains~\cite{chen2022vit,cha2021swad,bahri2022sam,kim2026reward}.
Random weight perturbation reaches flat minima by injecting noise into the parameters~\cite{li2024rwp}, and SAM~\cite{foret2021sam} by minimizing the worst-case loss within a radius $\rho$.
Since 3DGS is likewise an optimization over parameters, SAM applies to \cref{eq:update3dgs} directly,}
\begin{equation}
  \rev{\min_{\params}\ \max_{\|\pert\|\le\rho}\mathcal{L}_v(\params+\pert),\qquad \pert\approx\rho\,\frac{\gv}{\|\gv\|},}
  \label{eq:sam}
\end{equation}
\rev{where $\pert$ approximates the inner maximization by one ascent step along the clean gradient $\gv=\nabla_{\params}\mathcal{L}_v(\params)$, and $\params$ descends with the perturbed gradient $\tgv=\nabla_{\params}\mathcal{L}_v(\params+\pert)$.
We write $\gv_i$ and $\tgv_i$ for the parts of $\gv$ and $\tgv$ that belong to Gaussian $i$.
In 3DGS, noise on the Gaussian positions serves exploration in 3DGS-MCMC~\cite{kheradmand2024mcmc}, and two methods seek flat minima for sparse-view 3DGS.
FASR~\cite{yun2026flat} perturbs every Gaussian attribute along its gradient and, for the mean, rotation, and scale, sets the radius from the local image frequency,}
\begin{equation}
  \rev{\pert^{\text{FASR}}_i=\rho\,\tau_i\frac{\gv_i}{\|\gv_i\|},\qquad \tau_i=\frac{\gamma_i z_i}{f},}
  \label{eq:fasr}
\end{equation}
\rev{where $\gamma_i$ is the image scale at the projection of Gaussian $i$, which decreases with the local frequency, $z_i$ its rendered depth, and $f$ the focal length.
Unlike SAM, FASR keeps the clean loss in the objective for these geometric attributes,}
\begin{equation}
  \rev{\min_{\params}\ \frac{1-2\bar\gamma}{1-\bar\gamma}\,\mathcal{L}_v(\params)+\frac{\bar\gamma}{1-\bar\gamma}\,\mathcal{L}_v(\params+\pert^{\text{FASR}}),}
  \label{eq:update}
\end{equation}
\rev{with the weight set per Gaussian, $\bar\gamma_i=0.95\,\gamma_i/\gamma_{\max}$.
Gaussian $i$ thus descends along $\gv_i+\rev{\lambda}\dv_i$ with $\rev{\lambda=\bar\gamma/(1-\bar\gamma)}$ and the correction $\dv_i=\tgv_i-\gv_i$, which vanishes at a flat minimum.
SAP~\cite{seo2026sap} instead perturbs only the means $\boldsymbol{\mu}_i$, leaving the other parameters unchanged, with random noise that follows the covariance $\boldsymbol{\Sigma}_i$,}
\begin{equation}
  \rev{\min_{\params}\ \mathbb{E}\bigl[\mathcal{L}_v(\params+\pert^{\text{SAP}})\bigr],\quad \pert^{\text{SAP}}_i\sim\mathcal{N}(\mathbf{0},\,\sigma^2\boldsymbol{\Sigma}_i),}
  \label{eq:sap}
\end{equation}
\rev{with a scale $\sigma$, and periodically resets the scale, rotation, and higher-order SH coefficients.
Both perturb the Gaussians, but neither relates the perturbation direction to the gradients of the other training views.
\Cref{app:related} reviews related gradient decompositions~\cite{zhuang2022gsam,liu2022looksam,zhou2026gdags}.}

\section{\rev{Motivation}}
\label{sec:motivation}

\rev{A flat-minima optimizer perturbs each Gaussian and then updates it along the clean gradient plus the correction that this perturbation induces (\cref{eq:update}).
It thus chooses two directions: that of the perturbation, and that of the correction which enters the update.
FASR perturbs along the gradient of the current view and keeps the full correction, while SAP perturbs at random (\cref{sec:related}).
Under sparse views, however, the current view alone does not reveal what generalizes.}

\rev{\par\noindent\textbf{Hypothesis.} \emph{The gradients already carry this information: the gradients of the other training views indicate where a perturbation is useful, and the clean gradient indicates which part of the correction preserves the fit.}\par}
\rev{If so, two predictions follow, which \cref{sec:mot_pert,sec:mot_corr} test: a perturbation along the direction on which the views disagree should improve generalization, and a correction that opposes the clean gradient should impair the fit.}

\subsection{\rev{Direction of the perturbation}}
\label{sec:mot_pert}
\begin{table}[t]
\centering
\caption{\rev{\textbf{Direction of the perturbation} (LLFF, 3 views). Bottom: \rev{the FASR+SAP} perturbation redirected along the \rev{parallel or orthogonal component}. \rev{Noise: SAP noise.}}}
\label{tab:direction}\vspace{-0.5em}
\footnotesize
\setlength{\tabcolsep}{1.2pt}
\begin{tabular}{lccccc}
\toprule
Direction & \rev{Noise} & $\rho$ & PSNR$\uparrow$ & SSIM$\uparrow$ & LPIPS$\downarrow$ \\
\cmidrule(r{1pt}){1-1}\cmidrule(l{1pt}r{1pt}){2-2}\cmidrule(l{1pt}r{1pt}){3-3}\cmidrule(l{1pt}r{1pt}){4-6}
3DGS~\cite{kerbl2023gaussian} & \rev{\xmark} & -- & 19.74\sd{0.09} & 0.678\sd{0.001} & 0.214\sd{0.001} \\
\rev{+SAP~\cite{seo2026sap}} & \rev{\cmark} & -- & 20.16\sd{0.04} & 0.690\sd{0.001} & 0.207\sd{0.001} \\
\rev{+FASR~\cite{yun2026flat}} & \rev{\xmark} & 1 & 20.67\sd{0.06} & 0.715\sd{0.001} & 0.198\sd{0.001} \\
\rev{+FASR+SAP}~\cite{yun2026flat,seo2026sap} & \rev{\cmark} & 1 & 20.69\sd{0.06} & 0.715\sd{0.001} & 0.198\sd{0.001} \\
\cmidrule(r{1pt}){1-1}\cmidrule(l{1pt}r{1pt}){2-2}\cmidrule(l{1pt}r{1pt}){3-3}\cmidrule(l{1pt}r{1pt}){4-6}
\multirow{3}{*}{\rev{Parallel ($\gpar$)}} & \multirow{3}{*}{\rev{\cmark}} & 1 & 20.67\sd{0.06} & 0.710\sd{0.001} & 0.198\sd{0.001} \\
 & & 2 & 20.62\sd{0.12} & 0.706\sd{0.001} & 0.227\sd{0.001} \\
 & & 4 & 19.44\sd{0.04} & 0.606\sd{0.001} & 0.298\sd{0.001} \\
\cmidrule(r{1pt}){1-1}\cmidrule(l{1pt}r{1pt}){2-2}\cmidrule(l{1pt}r{1pt}){3-3}\cmidrule(l{1pt}r{1pt}){4-6}
\multirow{3}{*}{\rev{\textbf{Orthogonal ($\gperp$, Ours)}}} & \multirow{3}{*}{\rev{\cmark}} & 1 & 20.66\sd{0.03} & 0.713\sd{0.001} & \textbf{0.195}\sd{0.001} \\
 & & 2 & \cellcolor{ourgreen}\textbf{20.71}\sd{0.07} & \cellcolor{ourgreen}0.714\sd{0.001} & \cellcolor{ourgreen}\textbf{0.195}\sd{0.001} \\
 & & 4 & 20.70\sd{0.07} & \textbf{0.720}\sd{0.001} & 0.210\sd{0.001} \\
\bottomrule
\end{tabular}
\end{table}

\rev{To test the first prediction, we relate the gradient of the current view to those of the other training views.
Let $\gv^{(k)}_i=\nabla_{\boldsymbol{\mu}_i}\mathcal{L}_k(\params)$ be the gradient of the loss of training view $k$ with respect to the position of Gaussian $i$, and $\gv_i=\gv^{(v)}_i$ that of the current view $v$.
We decompose $\gv_i$ along the direction $\hat{\mathbf{a}}_i$ of the average gradient $\gbar_i$ of all training views,}
\begin{equation}
  \rev{\gbar_i=\frac{1}{|\mathcal{V}|}\sum_{k\in\mathcal{V}}\gv^{(k)}_i,\qquad \hat{\mathbf{a}}_i=\frac{\gbar_i}{\|\gbar_i\|},}
  \label{eq:consensus}
\end{equation}
\begin{equation}
  \rev{\mathbf{g}_{\parallel,i}=\langle\gv_i,\hat{\mathbf{a}}_i\rangle\,\hat{\mathbf{a}}_i,\qquad \mathbf{g}_{\perp,i}=\gv_i-\mathbf{g}_{\parallel,i}.}
  \label{eq:decomp}
\end{equation}
\rev{The parallel component moves the Gaussian in the direction shared with the other views, whereas the orthogonal component moves it in a direction that they do not share.}
\rev{This component is not a small residual: it carries about half of the gradient of the current view (\cref{tab:e2} in the appendix).}

\rev{\Cref{tab:direction} first shows that FASR and SAP each improve 3DGS, \rev{but their combination improves little over FASR alone}.}
\rev{Interestingly, the decomposition of \cref{eq:decomp} determines whether such a perturbation helps.
The bottom block of \cref{tab:direction} then replaces the direction of the FASR perturbation (\cref{eq:fasr}) by $\mathbf{g}_{\parallel,i}$ or $\mathbf{g}_{\perp,i}$ at the same radius.}
\rev{Perturbing along $\mathbf{g}_{\parallel,i}$ degrades the reconstruction as the radius grows, down to below plain 3DGS at $\rho=4$, whereas perturbing along $\mathbf{g}_{\perp,i}$ keeps the accuracy of FASR+SAP at every radius.
\rev{The same ordering holds without the noise of SAP (\cref{tab:direction_nosap}), so the direction decides whether the perturbation helps.}
Flatness thus helps where the views disagree, and not where they already agree.}
\rev{The noise, in turn, raises the PSNR of the orthogonal perturbation at every radius but not that of the parallel one, so \method keeps the orthogonal direction and adds the noise as a complement.}

\subsection{\rev{Direction of the correction}}
\label{sec:mot_corr}
\rev{To \rev{validate} the second prediction, we examine the update of FASR, which descends along $\gv_i+\lambda\dv_i$ for the position, rotation, and scale (\cref{eq:update}).
We split the correction into its projection onto the clean gradient and the remainder,}
\begin{equation}
  \rev{\gv_i+\lambda\dv_i=\underbrace{\gv_i+\lambda\dv_i^{\gv}}_{\text{reconstruction step}}+\underbrace{\lambda\,(\dv_i-\dv_i^{\gv})}_{\text{orthogonal to }\gv_i},}
  \label{eq:rescale}
\end{equation}
\rev{where $\dv_i^{\gv}=\frac{\langle\dv_i,\gv_i\rangle}{\|\gv_i\|^2}\,\gv_i$. Since the second term is orthogonal to $\gv_i$, only the first term, the reconstruction step, changes $\mathcal{L}_v$ to first order, and \cref{tab:e1} measures how often the correction works against this step during FASR training.
A correction that opposes the clean gradient ($\langle\dv_i,\gv_i\rangle<0$) shrinks this step, in one third to one half of the updates, and one whose opposing part outweighs the clean gradient ($\lambda\|\dv_i^{\gv}\|>\|\gv_i\|$) reverses it and increases $\mathcal{L}_v$, in $3$~to~$5\%$ of them.
The opposing part thus impairs the fit for a large share of the Gaussians, and more often for the rotation and scale than for the position, so we remove it from the correction of all three (\cref{sec:cau}).}
\begin{table}[t]
\centering\footnotesize
\caption{\rev{\textbf{Corrections that oppose the clean gradient} during FASR training (LLFF, 3 views): share of the per-Gaussian updates that shrink or reverse the reconstruction step.}}
\label{tab:e1}
\setlength{\tabcolsep}{5pt}
\begin{tabular}{lcc}
\toprule
Parameter & \rev{Shrunk} & \rev{Reversed} \\
 & \rev{$\langle\dv_i,\gv_i\rangle<0$} & \rev{$\lambda\|\dv_i^{\gv}\|>\|\gv_i\|$} \\
\cmidrule(r){1-1}\cmidrule(lr){2-2}\cmidrule(lr){3-3}
Position & 29.9\% & 3.2\% \\
Rotation & 48.3\% & 4.9\% \\
Scale & 42.0\% & 5.4\% \\
\bottomrule
\end{tabular}
\end{table}

\subsection{\rev{Overfitting of the color}}
\label{sec:mot_color}
\begin{figure}[t]
  \centering
  \begin{minipage}[t]{0.495\linewidth}\centering
    \includegraphics[width=\linewidth]{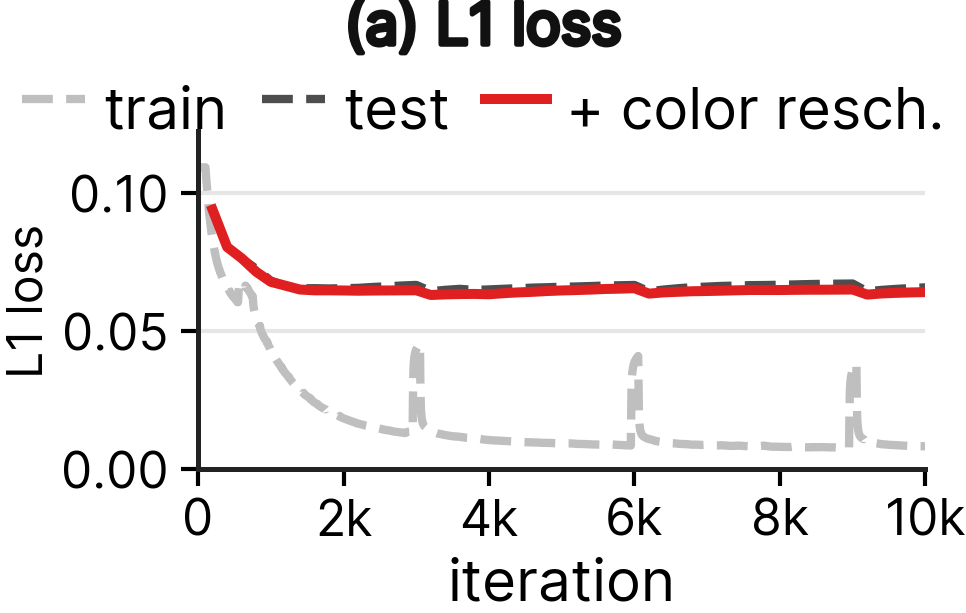}
  \end{minipage}\hfill
  \begin{minipage}[t]{0.495\linewidth}\centering
    \includegraphics[width=\linewidth]{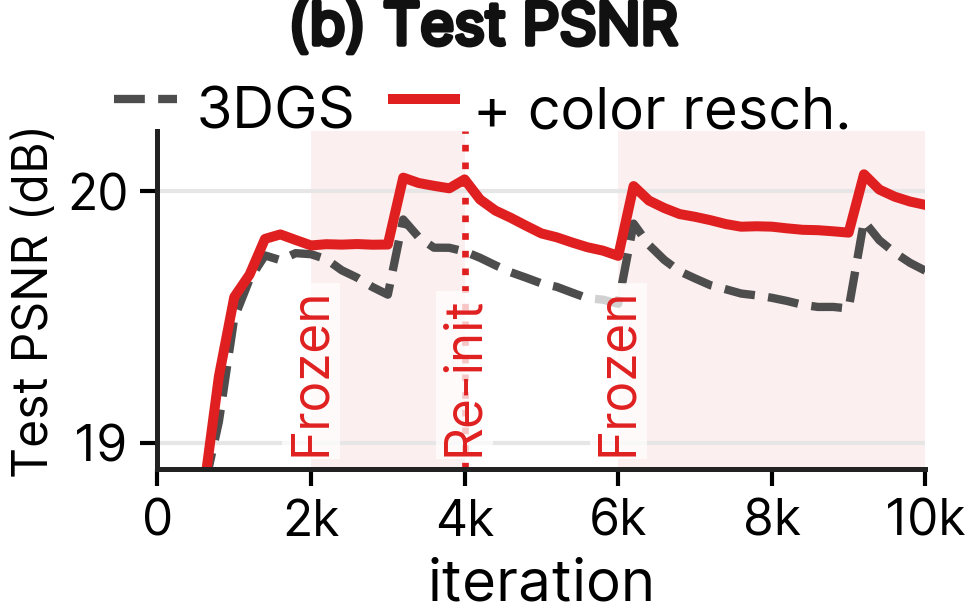}
  \end{minipage}
  \caption{\rev{\textbf{Generalization gap} (LLFF, 3 views).} \rev{Train and test loss of 3DGS and test loss with color rescheduling; test PSNR without and with it. Single run; \cref{tab:shr}: mean over 5 runs.}}
  \label{fig:sh_curves}
\end{figure}
\rev{The two directions above concern the geometry of the Gaussians, not their view-dependent color.
To examine whether the color overfits on its own, \cref{fig:sh_curves} tracks the training and test losses of plain 3DGS with three views.
The test loss turns upward after about 3k iterations while the training loss keeps falling (\cref{fig:sh_curves}a), so the color fitted after that point explains the training views alone.
The test PSNR thus peaks at 19.89\,dB around 3k iterations and ends at 19.68\,dB (\cref{fig:sh_curves}b).
\Cref{sec:method} remedies each observation in turn.}

\section{\rev{Main Contribution: DecoFlat}}
\label{sec:method}

\rev{Based on these observations, we propose \method, which \rev{starts from FASR+SAP and revises it} in three respects, each following one observation of \cref{sec:motivation} (\cref{fig:overview}, \cref{alg:decoflat}).
It perturbs the Gaussians along the orthogonal component (\cref{sec:vas}), removes the opposing part of the correction (\cref{sec:cau}), and reschedules the color (\cref{sec:shr}).}

\begin{algorithm}[t]
\caption{\rev{Pseudocode for \method}}
\label{alg:decoflat}
\footnotesize\revcolor
\begin{algorithmic}[1]
\Require views $\mathcal{V}$, cache $C$, warm-up $T_0$
\For{$t=1,\dots,T$}
  \State Sample $v\in\mathcal{V}$;\ \ $\gv\leftarrow\nabla_{\params}\mathcal{L}_v(\params)$;\ \ $C[v]\leftarrow\gv^{(v)}$ \Comment{clean pass}
  \If{$t\ge T_0$}
    \State $\pert^{\text{ours}}\leftarrow$ \cref{eq:vas}, with $\gbar$ of \cref{eq:cache_mean}
    \State $\tgv\leftarrow\nabla_{\params}\mathcal{L}_v(\params+\pert^{\text{ours}})$ \Comment{perturbed pass}
    \State $\dv\leftarrow\tgv-\gv$;\ \ $\uv\leftarrow$ \cref{eq:cau}
  \Else
    \State $\uv\leftarrow\gv$
  \EndIf
  \State \rev{Update $\params$ with $\uv$; gate the SH step by} \cref{eq:shr}
\EndFor
\end{algorithmic}
\end{algorithm}

\subsection{\rev{Orthogonal perturbation}}
\label{sec:vas}
\rev{Flatness helps along the direction on which the training views disagree, and not along the one on which they agree (\cref{sec:mot_pert}).}
\rev{To seek flatness only where it helps, we perturb each Gaussian along the orthogonal component,}
\begin{equation}
  \rev{\pert^{\text{ours}}_i=\rho\,\tau_i\,\frac{\mathbf{g}_{\perp,i}}{\|\mathbf{g}_{\perp,i}\|}}\rev{\;+\;\boldsymbol{\epsilon}_i,\qquad \boldsymbol{\epsilon}_i\sim\mathcal{N}(\mathbf{0},\sigma^{2}\boldsymbol{\Sigma}_i),}
  \label{eq:vas}
\end{equation}
\rev{applied to the mean $\boldsymbol{\mu}_i$, with $\mathbf{g}_{\perp,i}$ of \cref{eq:decomp} and $\tau_i$ of \cref{eq:fasr}\rev{, and Gaussian noise $\boldsymbol{\epsilon}_i$ shaped by the covariance $\boldsymbol{\Sigma}_i$ with scale $\sigma$.}}

\rev{Since the orthogonal term has no component along the average gradient, it leaves the average training loss unchanged to first order, so the resulting correction responds only to the sharpness that the current view does not share with the others.}

\subsection{\rev{Conflict-aware correction}}
\label{sec:cau}
\rev{A correction that opposes the clean gradient shrinks or even reverses the reconstruction step, as analyzed in \cref{sec:mot_corr}.}
\rev{To preserve the fit to the training views, we drop its projection $\dv_i^{\gv}$ in that case only,}
\begin{equation}
  \rev{\uv_i=\begin{cases}\gv_i+\lambda\dv_i, & \langle\dv_i,\gv_i\rangle\ge0,\\ \gv_i+\lambda\,(\dv_i-\dv_i^{\gv}), & \text{otherwise},\end{cases}}
  \label{eq:cau}
\end{equation}

\rev{where $\uv_i$ is the direction passed to the optimizer for the position, rotation, and scale of Gaussian $i$.
\rev{When the correction agrees with the clean gradient, it is kept in full; otherwise, only its opposing part is removed, so the reconstruction step stays $\gv_i$ and the rest of the correction is kept, as \cref{prop:cau} states.}}
\begin{proposition}
\label{prop:cau}
\rev{For every Gaussian, $\langle\uv_i,\gv_i\rangle=\|\gv_i\|^2+\lambda\max(\langle\dv_i,\gv_i\rangle,0)\ge\|\gv_i\|^2$, and the component of $\uv_i$ orthogonal to $\gv_i$ equals that of \rev{$\gv_i+\lambda\dv_i$}, $\lambda(\dv_i-\dv_i^{\gv})$.}
\end{proposition}
\rev{The update therefore descends the reconstruction loss at least as fast as plain gradient descent to first order, while keeping the part of the correction that moves the Gaussian toward a flatter region (proof in \cref{app:proofs}).
\rev{This mirrors GSAM~\cite{zhuang2022gsam} with the roles exchanged: we decompose the correction against the clean gradient, rather than the clean gradient against the perturbed one.}}

\begin{table*}[t]
\centering
\caption{\rev{\textbf{Quantitative comparison of 3DGS and sparse-view methods with and without flat-minima optimizers}} (LLFF, 3 and 6 views; Mip-NeRF\,360, 12 views)\rev{. Mean and standard deviation over 5 runs. Bold: best per block.}}
\label{tab:main}\label{tab:plugin_llff}\vspace{-0.5em}
\scriptsize
\setlength{\tabcolsep}{2.2pt}
\begin{tabular*}{0.95\textwidth}{@{\extracolsep{\fill}}llccccccccc}
\toprule
Baseline & Method & \multicolumn{3}{c}{LLFF, 3 views} & \multicolumn{3}{c}{LLFF, 6 views} & \multicolumn{3}{c}{Mip-NeRF\,360, 12 views} \\
\cmidrule(lr){3-5}\cmidrule(lr){6-8}\cmidrule(lr){9-11}
 & & PSNR$\uparrow$ & SSIM$\uparrow$ & LPIPS$\downarrow$ & PSNR$\uparrow$ & SSIM$\uparrow$ & LPIPS$\downarrow$ & PSNR$\uparrow$ & SSIM$\uparrow$ & LPIPS$\downarrow$ \\
\cmidrule(r){1-1}\cmidrule(r){2-2}\cmidrule(lr){3-5}\cmidrule(lr){6-8}\cmidrule(lr){9-11}
\multirow{4}{*}{\shortstack[l]{3DGS\\\cite{kerbl2023gaussian}}}
 & Baseline & 19.74\sd{0.09} & 0.678\sd{0.001} & 0.214\sd{0.001} & 23.23\sd{0.08} & 0.788\sd{0.002} & 0.147\sd{0.001} & 19.14\sd{0.02} & 0.559\sd{0.001} & 0.365\sd{0.001} \\
 & +SAP~\cite{seo2026sap} & 20.16\sd{0.04} & 0.690\sd{0.001} & 0.207\sd{0.001} & 23.39\sd{0.07} & 0.794\sd{0.002} & 0.145\sd{0.001} & 19.29\sd{0.06} & 0.565\sd{0.002} & 0.358\sd{0.001} \\
 & +FASR~\cite{yun2026flat} & 20.67\sd{0.06} & 0.715\sd{0.001} & 0.198\sd{0.001} & 24.05\sd{0.03} & 0.811\sd{0.001} & 0.143\sd{0.001} & 19.38\sd{0.08} & 0.570\sd{0.001} & \textbf{0.352}\sd{0.001} \\
\cmidrule(r){2-2}\cmidrule(lr){3-5}\cmidrule(lr){6-8}\cmidrule(lr){9-11}
 & \cellcolor{ourgreen}\textbf{\rev{+\method}~\rev{(Ours)}} & \cellcolor{ourgreen}\textbf{20.84}\sd{0.08} & \cellcolor{ourgreen}\textbf{0.716}\sd{0.001} & \cellcolor{ourgreen}\textbf{0.195}\sd{0.001} & \cellcolor{ourgreen}\textbf{24.11}\sd{0.06} & \cellcolor{ourgreen}\textbf{0.815}\sd{0.001} & \cellcolor{ourgreen}\textbf{0.140}\sd{0.001} & \cellcolor{ourgreen}\textbf{19.62}\sd{0.03} & \cellcolor{ourgreen}\textbf{0.577}\sd{0.000} & \cellcolor{ourgreen}0.354\sd{0.001} \\
\midrule
\multirow{4}{*}{\shortstack[l]{CoR-GS\\\cite{zhang2024corgs}}}
 & Baseline &  20.16\sd{0.02} & 0.702\sd{0.001} & 0.202\sd{0.000} &  23.44\sd{0.05} & 0.793\sd{0.001} & \textbf{0.150}\sd{0.001} & 20.01\sd{0.09} & 0.592\sd{0.002} & 0.376\sd{0.001} \\
 & +SAP~\cite{seo2026sap} &  20.41\sd{0.01} & 0.706\sd{0.001} & \textbf{0.201}\sd{0.000} &  23.48\sd{0.04} & 0.794\sd{0.001} & 0.151\sd{0.001} & 20.08\sd{0.05} & \textbf{0.596}\sd{0.001} & \textbf{0.374}\sd{0.001} \\
 & +FASR~\cite{yun2026flat} &  20.87\sd{0.14} & 0.726\sd{0.001} & 0.210\sd{0.001} &  24.02\sd{0.05} & 0.806\sd{0.001} & 0.169\sd{0.001} & 20.16\sd{0.04} & 0.586\sd{0.002} & 0.394\sd{0.001} \\
\cmidrule(r){2-2}\cmidrule(lr){3-5}\cmidrule(lr){6-8}\cmidrule(lr){9-11}
 & \cellcolor{ourgreen}\textbf{\rev{+\method}~\rev{(Ours)}} & \cellcolor{ourgreen}\textbf{20.97}\sd{0.21} & \cellcolor{ourgreen}\textbf{0.731}\sd{0.002} & \cellcolor{ourgreen}\textbf{0.201}\sd{0.001} & \cellcolor{ourgreen}\textbf{24.07}\sd{0.06} & \cellcolor{ourgreen}\textbf{0.810}\sd{0.001} & \cellcolor{ourgreen}0.166\sd{0.001} & \cellcolor{ourgreen}\textbf{20.30}\sd{0.07} & \cellcolor{ourgreen}0.589\sd{0.001} & \cellcolor{ourgreen}0.402\sd{0.001} \\
\midrule
\multirow{4}{*}{\shortstack[l]{SE-GS\\\cite{segs2025}}}
 & Baseline &  20.71\sd{0.05} & 0.720\sd{0.001} & 0.186\sd{0.001} &  24.00\sd{0.06} & 0.810\sd{0.001} & \textbf{0.131}\sd{0.001} & 20.38\sd{0.03} & 0.605\sd{0.001} & 0.377\sd{0.001} \\
 & +SAP~\cite{seo2026sap} &  20.80\sd{0.04} & 0.723\sd{0.002} & \textbf{0.184}\sd{0.001} &  23.99\sd{0.04} & 0.810\sd{0.001} & 0.133\sd{0.001} & 20.41\sd{0.08} & \textbf{0.607}\sd{0.002} & \textbf{0.376}\sd{0.001} \\
 & +FASR~\cite{yun2026flat} &  20.86\sd{0.23} & 0.732\sd{0.003} & 0.202\sd{0.004} &  24.39\sd{0.02} & 0.818\sd{0.001} & 0.152\sd{0.001} & 20.42\sd{0.06} & 0.599\sd{0.002} & 0.397\sd{0.002} \\
\cmidrule(r){2-2}\cmidrule(lr){3-5}\cmidrule(lr){6-8}\cmidrule(lr){9-11}
 & \cellcolor{ourgreen}\textbf{\rev{+\method}~\rev{(Ours)}} & \cellcolor{ourgreen}\textbf{21.15}\sd{0.16} & \cellcolor{ourgreen}\textbf{0.742}\sd{0.003} & \cellcolor{ourgreen}0.190\sd{0.003} & \cellcolor{ourgreen}\textbf{24.41}\sd{0.07} & \cellcolor{ourgreen}\textbf{0.824}\sd{0.001} & \cellcolor{ourgreen}0.149\sd{0.001} & \cellcolor{ourgreen}\textbf{20.54}\sd{0.06} & \cellcolor{ourgreen}0.600\sd{0.002} & \cellcolor{ourgreen}0.404\sd{0.001} \\
\midrule
\multirow{4}{*}{\shortstack[l]{DropGaussian\\\cite{park2025dropgaussian}}}
 & Baseline &  20.51\sd{0.06} & 0.709\sd{0.001} & 0.201\sd{0.001} &  23.62\sd{0.03} & 0.801\sd{0.001} & 0.147\sd{0.001} & 19.81\sd{0.06} & 0.583\sd{0.001} & 0.359\sd{0.001} \\
 & +SAP~\cite{seo2026sap} &  20.49\sd{0.09} & 0.710\sd{0.001} & 0.201\sd{0.001} &  23.77\sd{0.05} & 0.803\sd{0.001} & 0.145\sd{0.001} & 19.80\sd{0.04} & 0.584\sd{0.001} & 0.356\sd{0.001} \\
 & +FASR~\cite{yun2026flat} &  20.74\sd{0.03} & 0.722\sd{0.001} & 0.195\sd{0.001} &  24.03\sd{0.08} & 0.814\sd{0.001} & 0.144\sd{0.001} & 19.81\sd{0.08} & 0.584\sd{0.002} & \textbf{0.352}\sd{0.001} \\
\cmidrule(r){2-2}\cmidrule(lr){3-5}\cmidrule(lr){6-8}\cmidrule(lr){9-11}
 & \cellcolor{ourgreen}\textbf{\rev{+\method}~\rev{(Ours)}} & \cellcolor{ourgreen}\textbf{20.80}\sd{0.18} & \cellcolor{ourgreen}\textbf{0.724}\sd{0.002} & \cellcolor{ourgreen}\textbf{0.192}\sd{0.001} & \cellcolor{ourgreen}\textbf{24.15}\sd{0.10} & \cellcolor{ourgreen}\textbf{0.817}\sd{0.001} & \cellcolor{ourgreen}\textbf{0.143}\sd{0.001} & \cellcolor{ourgreen}\textbf{19.96}\sd{0.03} & \cellcolor{ourgreen}\textbf{0.591}\sd{0.001} & \cellcolor{ourgreen}0.354\sd{0.001} \\
\midrule
\multirow{4}{*}{\shortstack[l]{Scaffold-GS\\\cite{lu2024scaffold}}}
 & Baseline &  18.83\sd{0.04} & 0.652\sd{0.001} & 0.245\sd{0.001} &  22.43\sd{0.04} & 0.768\sd{0.001} & 0.166\sd{0.000} & 18.57\sd{0.06} & 0.535\sd{0.002} & 0.402\sd{0.001} \\
 & +SAP~\cite{seo2026sap} &  19.18\sd{0.07} & 0.672\sd{0.001} & 0.226\sd{0.001} &  22.63\sd{0.05} & 0.778\sd{0.001} & 0.156\sd{0.001} & 18.68\sd{0.05} & 0.545\sd{0.000} & 0.392\sd{0.001} \\
 & +FASR~\cite{yun2026flat} &  19.49\sd{0.08} & 0.688\sd{0.001} & 0.221\sd{0.001} &  22.91\sd{0.08} & 0.788\sd{0.001} & 0.157\sd{0.002} & 18.80\sd{0.04} & 0.554\sd{0.001} & \textbf{0.391}\sd{0.000} \\
\cmidrule(r){2-2}\cmidrule(lr){3-5}\cmidrule(lr){6-8}\cmidrule(lr){9-11}
 & \cellcolor{ourgreen}\textbf{\rev{+\method}~\rev{(Ours)}} & \cellcolor{ourgreen}\textbf{19.92}\sd{0.03} & \cellcolor{ourgreen}\textbf{0.703}\sd{0.001} & \cellcolor{ourgreen}\textbf{0.205}\sd{0.001} & \cellcolor{ourgreen}\textbf{23.35}\sd{0.05} & \cellcolor{ourgreen}\textbf{0.799}\sd{0.001} & \cellcolor{ourgreen}\textbf{0.146}\sd{0.001} & \cellcolor{ourgreen}\textbf{19.00}\sd{0.07} & \cellcolor{ourgreen}\textbf{0.559}\sd{0.001} & \cellcolor{ourgreen}\textbf{0.391}\sd{0.001} \\
\midrule
\multirow{4}{*}{\shortstack[l]{DNGaussian\\\cite{li2024dngaussian}}}
 & Baseline &  18.80\sd{0.05} & 0.592\sd{0.004} & 0.302\sd{0.005} &  22.02\sd{0.07} & 0.717\sd{0.002} & \textbf{0.249}\sd{0.003} & \textbf{16.59}\sd{0.09} & \textbf{0.391}\sd{0.003} & \textbf{0.563}\sd{0.003} \\
 & +SAP~\cite{seo2026sap} &  18.94\sd{0.08} & 0.602\sd{0.003} & 0.293\sd{0.002} &  21.96\sd{0.04} & 0.717\sd{0.002} & 0.252\sd{0.001} & 15.67\sd{0.06} & 0.351\sd{0.004} & 0.598\sd{0.004} \\
 & +FASR~\cite{yun2026flat} &  18.86\sd{0.22} & 0.607\sd{0.006} & 0.297\sd{0.005} &  21.71\sd{0.11} & 0.713\sd{0.001} & 0.259\sd{0.001} & 16.25\sd{0.10} & 0.384\sd{0.005} & 0.586\sd{0.006} \\
\cmidrule(r){2-2}\cmidrule(lr){3-5}\cmidrule(lr){6-8}\cmidrule(lr){9-11}
 & \cellcolor{ourgreen}\textbf{\rev{+\method}~\rev{(Ours)}} & \cellcolor{ourgreen}\textbf{19.26}\sd{0.06} & \cellcolor{ourgreen}\textbf{0.614}\sd{0.003} & \cellcolor{ourgreen}\textbf{0.290}\sd{0.002} & \cellcolor{ourgreen}\textbf{22.11}\sd{0.05} & \cellcolor{ourgreen}\textbf{0.722}\sd{0.002} & \cellcolor{ourgreen}0.254\sd{0.001} & \cellcolor{ourgreen}16.39\sd{0.12} & \cellcolor{ourgreen}0.387\sd{0.006} & \cellcolor{ourgreen}0.579\sd{0.006} \\
\midrule
\multirow{4}{*}{\shortstack[l]{FSGS\\\cite{zhu2024fsgs}}}
 & Baseline & 20.48\sd{0.02} & 0.699\sd{0.001} & \textbf{0.205}\sd{0.001} & 24.12\sd{0.05} & \textbf{0.819}\sd{0.002} & 0.126\sd{0.001} & 19.23\sd{0.09} & 0.554\sd{0.002} & 0.375\sd{0.001} \\
 & +SAP~\cite{seo2026sap} & 20.49\sd{0.10} & 0.700\sd{0.002} & \textbf{0.205}\sd{0.002} & 24.14\sd{0.06} & 0.817\sd{0.001} & \textbf{0.124}\sd{0.001} & 19.22\sd{0.06} & 0.553\sd{0.001} & 0.374\sd{0.001} \\
 & +FASR~\cite{yun2026flat} & 20.61\sd{0.06} & 0.703\sd{0.001} & 0.216\sd{0.001} & 24.19\sd{0.08} & 0.815\sd{0.002} & 0.148\sd{0.001} & 19.41\sd{0.12} & 0.558\sd{0.003} & \textbf{0.371}\sd{0.002} \\
\cmidrule(r){2-2}\cmidrule(lr){3-5}\cmidrule(lr){6-8}\cmidrule(lr){9-11}
 & \cellcolor{ourgreen}\textbf{\rev{+\method}~\rev{(Ours)}} & \cellcolor{ourgreen}\textbf{20.78}\sd{0.06} & \cellcolor{ourgreen}\textbf{0.710}\sd{0.002} & \cellcolor{ourgreen}0.209\sd{0.001} & \cellcolor{ourgreen}\textbf{24.29}\sd{0.08} & \cellcolor{ourgreen}\textbf{0.819}\sd{0.001} & \cellcolor{ourgreen}0.134\sd{0.001} & \cellcolor{ourgreen}\textbf{19.53}\sd{0.09} & \cellcolor{ourgreen}\textbf{0.560}\sd{0.002} & \cellcolor{ourgreen}0.376\sd{0.001} \\
\bottomrule
\end{tabular*}
\end{table*}

\subsection{\rev{Color rescheduling}}
\label{sec:shr}
\rev{As observed in \cref{sec:mot_color}, the color fitted after about 3k iterations explains the training views alone.}
\rev{To discard this part while keeping the early color, we freeze the SH coefficients $\mathbf{c}_i$ at iteration $T_{\mathrm f}$, re-initialize their higher-degree part at $T_{\mathrm r}$ and refit them on the settled geometry, and freeze them again from $T_{\mathrm e}$ on.
Equivalently, the learning rate of $\mathbf{c}_i$ is multiplied by the gate}
\begin{equation}
  \rev{m(t)=\mathbf{1}[t<T_{\mathrm f}]+\mathbf{1}[T_{\mathrm r}\le t<T_{\mathrm e}],}
  \label{eq:shr}
\end{equation}
\rev{where $\mathbf{1}[\cdot]$ is the indicator function, and the \rev{optimizer} moments of the re-initialized coefficients are reset (\cref{fig:overview}c).}

\begin{figure}[t]
  \centering
  \includegraphics[width=1.0\linewidth]{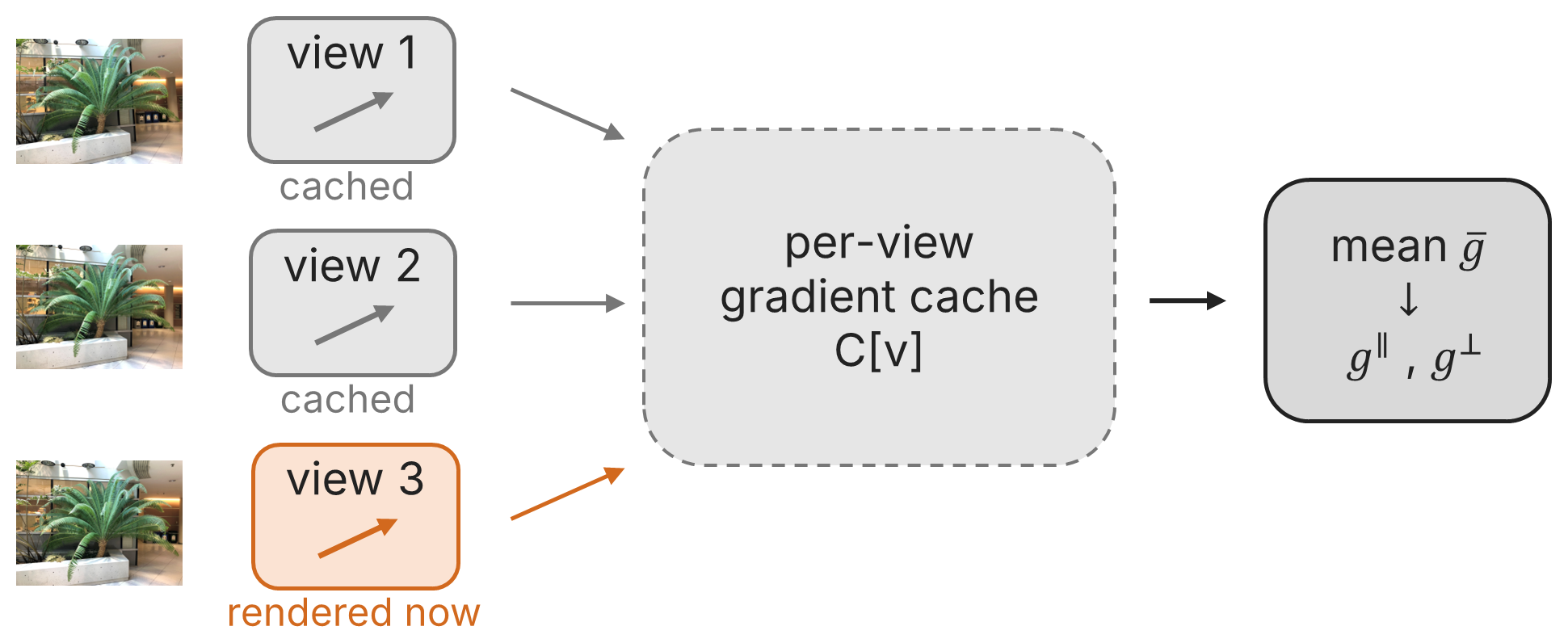}
  \caption{\rev{\textbf{Gradient caching.} Only the current view is rendered; the gradients of the other views are read from the per-view cache $C$, and their average $\gbar$ gives $\gpar$ and $\gperp$.}}
  \label{fig:cache}
\end{figure}
\subsection{\rev{Practical implementation: Gradient Caching}}
\label{sec:cache}
\rev{Computing $\gbar_i$ of \cref{eq:consensus} exactly requires a clean pass for every training view at every iteration, which multiplies the training cost by $|\mathcal{V}|$.
Since the gradient of a view changes little between two visits, we instead keep a cache $C$ that stores, for each view $k$, the position gradients $C[k]_i=\gv^{(k)}_i(\params^{(t_k)})$ of its most recent clean pass at iteration $t_k\le t$.
The clean pass of the current view $v$ overwrites $C[v]$ at no extra cost, and $\gbar_i$ is approximated by}
\begin{equation}
  \rev{\gbar_i\approx\frac{1}{|\mathcal{V}|}\sum_{k\in\mathcal{V}}C[k]_i.}
  \label{eq:cache_mean}
\end{equation}
\rev{Only the entry of the current view is exact, yet with 3 LLFF views the approximation agrees with the exact average at a median per-Gaussian cosine of 0.99 (\cref{app:scope}).
The cache is refreshed after densification or pruning, which change the set of Gaussians.
\rev{\Cref{fig:cache} illustrates the cache.}}

\section{\rev{Experiments}}
\label{sec:exp}

\subsection{\rev{Setup}}
\label{sec:exp_setup}
\rev{\noindent\textbf{Datasets and metrics.}
We follow the sparse-view protocol of FASR~\cite{yun2026flat} on LLFF~\cite{mildenhall2019llff} with 3 and 6 input views \rev{(8 scenes)} and Mip-NeRF~360~\cite{barron2022mipnerf360} with 12 input views \rev{(7 scenes)}, the 8-view Blender~\cite{mildenhall2020nerf} setting of FSGS~\cite{zhu2024fsgs} \rev{(8 scenes)}\rev{, and the 3-view DTU~\cite{jensen2014dtu} setting of RegNeRF~\cite{niemeyer2022regnerf} (15 scans)}\rev{, at the resolutions and with the view splits of the respective baselines}.
We report PSNR, SSIM\rev{~\cite{wang2004ssim}}, and LPIPS~\cite{zhang2018lpips}, averaged over scenes\rev{, as the mean and standard deviation over 5 runs, with paired Wilcoxon signed-rank tests\rev{~\cite{wilcoxon1945}} for significance.}}

\rev{\noindent\textbf{Baselines.}
On the 3DGS backbone, we compare \method with the flat-minima optimizers SAP~\cite{seo2026sap} and FASR~\cite{yun2026flat} in the official FASR code under identical schedules.
As a plug-in, we add each optimizer to the official code of CoR-GS~\cite{zhang2024corgs}, SE-GS~\cite{segs2025}, DropGaussian~\cite{park2025dropgaussian}, Scaffold-GS~\cite{lu2024scaffold}, DNGaussian~\cite{li2024dngaussian}\rev{, and FSGS~\cite{zhu2024fsgs}}, without changing their losses, priors, or schedules.
}

\rev{\noindent\textbf{Implementation details.}
All flat-minima optimizers share the schedule and hyperparameters of FASR\rev{,} start perturbing after a warm-up of $T_0=3$k iterations\rev{, and pass their update to Adam~\cite{kingma2015adam} as in 3DGS}.
\method uses $\rho=2$, the best radius in \cref{tab:direction}, and $(T_{\mathrm f},T_{\mathrm r},T_{\mathrm e})=(2\text{k},4\text{k},6\text{k})$ for color rescheduling.
DNGaussian, which trains for 6k iterations, and Scaffold-GS, which has no SH color, use only the orthogonal perturbation and the conflict-aware correction.
All experiments run on a single NVIDIA A100 GPU, and per-code details are given in \cref{app:details}.}

\subsection{\rev{Main results}}
\label{sec:exp_main}\label{sec:exp_plugin}
\begin{figure*}[!t]
  \centering
  \includegraphics[width=1.0\linewidth]{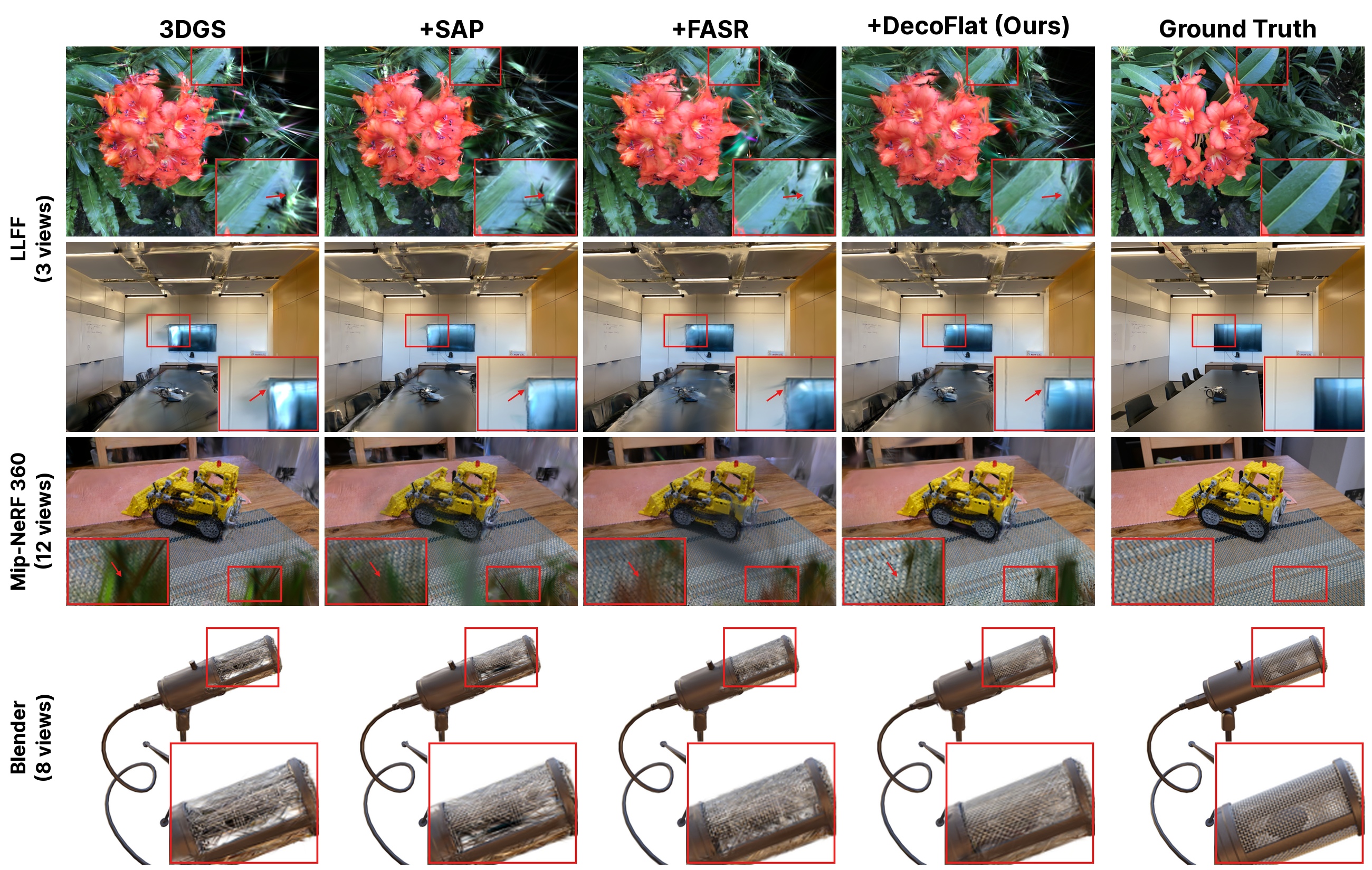}
  \caption{\rev{\textbf{Qualitative comparison on the 3DGS backbone} (LLFF, 3 views; Mip-NeRF\,360, 12 views; Blender, 8 views).}}
  \label{fig:qual}
\end{figure*}
\rev{\noindent\textbf{Quantitative results.}}
\rev{To verify that \method improves generalization both on its own and as a plug-in, \cref{tab:main} adds each flat-minima optimizer to 3DGS and to six sparse-view methods that differ in their supervision (a depth prior in DNGaussian and FSGS), regularization (co-training in CoR-GS, self-ensembling in SE-GS, dropout in DropGaussian), and representation (anchors in Scaffold-GS).
\rev{\method is the only optimizer that improves every method on LLFF, with both 3 and 6 views, and it achieves the best PSNR in every block, whereas SAP and FASR each degrade at least one method.
Its gain over FASR is largest on the anchor-based Scaffold-GS and the depth-supervised DNGaussian, by about 0.4\,dB, and on LLFF it keeps the LPIPS lower wherever FASR raises it.
The same holds on the unbounded scenes of Mip-NeRF\,360, where \method achieves the best PSNR for six of the seven methods; the exception is DNGaussian, which no optimizer improves.
\method thus acts as a general optimizer for sparse-view 3DGS rather than as a refinement of one pipeline.}}

\rev{\noindent\textbf{Qualitative results.}}
\rev{In \cref{fig:qual}, 3DGS places floaters in front of the surfaces to explain the training views, and SAP and FASR thin them out but leave holes and blurred edges; \method removes them while keeping the surfaces, consistent with seeking flatness only where the views disagree.}

\subsection{\rev{Results on additional datasets}}
\label{sec:exp_more}
\begin{table}[t]
\centering
\caption{\rev{\textbf{Plug-in results on Blender (8 views) and DTU (3 views)}. Bold: best per block.} \rev{Protocol in \cref{app:results}.}}
\label{tab:plugin_blender_dtu}\vspace{-0.5em}
\scriptsize
\setlength{\tabcolsep}{1.2pt}
\resizebox{0.95\linewidth}{!}{\begin{tabular}{llcccccc}
\toprule
Baseline & Method & \multicolumn{3}{c}{Blender, 8 views} & \multicolumn{3}{c}{DTU, 3 views} \\
\cmidrule(lr){3-5}\cmidrule(lr){6-8}
 & & PSNR$\uparrow$ & SSIM$\uparrow$ & LPIPS$\downarrow$ & PSNR$\uparrow$ & SSIM$\uparrow$ & LPIPS$\downarrow$ \\
\cmidrule(r){1-1}\cmidrule(r){2-2}\cmidrule(lr){3-5}\cmidrule(lr){6-8}
\multirow{4}{*}{\shortstack[l]{3DGS\\\cite{kerbl2023gaussian}}}
 & Baseline & 22.74 & 0.873 & 0.109 & 16.33 & 0.817 & 0.160 \\
 & +SAP~\cite{seo2026sap} & 23.03 & 0.877 & 0.106 & 17.15 & 0.829 & 0.147 \\
 & +FASR~\cite{yun2026flat} & 23.07 & 0.879 & \textbf{0.104} & 17.07 & 0.832 & 0.147 \\
\cmidrule(r){2-2}\cmidrule(lr){3-5}\cmidrule(lr){6-8}
 & \cellcolor{ourgreen}\textbf{\rev{+\method}~\rev{(Ours)}} & \cellcolor{ourgreen}\textbf{23.22} & \cellcolor{ourgreen}\textbf{0.882} & \cellcolor{ourgreen}0.106 & \cellcolor{ourgreen}\textbf{17.21} & \cellcolor{ourgreen}\textbf{0.833} & \cellcolor{ourgreen}\textbf{0.146} \\
\midrule
\multirow{4}{*}{\shortstack[l]{CoR-GS\\\cite{zhang2024corgs}}}
 & Baseline & 23.95 & 0.891 & 0.094 & 19.21 & 0.853 & 0.124 \\
 & +SAP~\cite{seo2026sap} & 24.17 & \textbf{0.893} & \textbf{0.093} & 19.67 & 0.854 & \textbf{0.120} \\
 & +FASR~\cite{yun2026flat} & 24.20 & \textbf{0.893} & 0.094 & \textbf{19.93} & 0.861 & 0.127 \\
\cmidrule(r){2-2}\cmidrule(lr){3-5}\cmidrule(lr){6-8}
 & \cellcolor{ourgreen}\textbf{\rev{+\method}~\rev{(Ours)}} & \cellcolor{ourgreen}\textbf{24.23} & \cellcolor{ourgreen}0.892 & \cellcolor{ourgreen}\textbf{0.093} & \cellcolor{ourgreen}19.73 & \cellcolor{ourgreen}\textbf{0.862} & \cellcolor{ourgreen}0.129 \\
\midrule
\multirow{4}{*}{\shortstack[l]{DNGaussian\\\cite{li2024dngaussian}}}
 & Baseline & 24.18 & 0.889 & 0.088 & 18.78 & 0.790 & 0.167 \\
 & +SAP~\cite{seo2026sap} & 24.40 & 0.892 & \textbf{0.086} & 18.65 & 0.803 & 0.157 \\
 & +FASR~\cite{yun2026flat} & 24.51 & \textbf{0.893} & 0.088 & 17.42 & 0.801 & 0.164 \\
\cmidrule(r){2-2}\cmidrule(lr){3-5}\cmidrule(lr){6-8}
 & \cellcolor{ourgreen}\textbf{\rev{+\method}~\rev{(Ours)}} & \cellcolor{ourgreen}\textbf{24.54} & \cellcolor{ourgreen}\textbf{0.893} & \cellcolor{ourgreen}0.088 & \cellcolor{ourgreen}\textbf{19.06} & \cellcolor{ourgreen}\textbf{0.822} & \cellcolor{ourgreen}\textbf{0.142} \\
\bottomrule
\end{tabular}}
\end{table}

\rev{To verify that \method generalizes across scene types, \cref{tab:plugin_blender_dtu} adds the optimizers to 3DGS, CoR-GS, and DNGaussian on the synthetic Blender and the object-centric DTU.
\method achieves the best PSNR in five of the six blocks, so its gain holds across scene types and backbones; the exception is CoR-GS on DTU, where FASR is higher.}

\section{\rev{Analysis}}
\label{sec:analysis}

\label{sec:exp_ablation}
\begin{table}[!t]
\centering
\caption{\rev{\textbf{Ablation of the three components} (LLFF, 3 views). Pert.: orthogonal perturbation; Corr.: conflict-aware correction; Color: color rescheduling. Bold: best.}}
\label{tab:ablation}\vspace{-0.5em}
\footnotesize
\setlength{\tabcolsep}{3.5pt}
\begin{tabular}{ccc ccc}
\toprule
Pert. & Corr. & Color & PSNR$\uparrow$ & SSIM$\uparrow$ & LPIPS$\downarrow$ \\
\cmidrule(r){1-3}\cmidrule(lr){4-6}
\xmark & \xmark & \xmark & 20.69\sd{0.06} & 0.715\sd{0.001} & 0.198\sd{0.001} \\
\cmark & \xmark & \xmark & 20.71\sd{0.07} & 0.714\sd{0.001} & \textbf{0.195}\sd{0.001} \\
\xmark & \cmark & \xmark & 20.75\sd{0.08} & 0.717\sd{0.001} & 0.197\sd{0.001} \\
\xmark & \xmark & \cmark & 20.74\sd{0.14} & \textbf{0.720}\sd{0.001} & 0.199\sd{0.001} \\
\cmark & \cmark & \xmark & 20.65\sd{0.07} & 0.712\sd{0.002} & \textbf{0.195}\sd{0.001} \\
\cmark & \xmark & \cmark & 20.79\sd{0.11} & 0.718\sd{0.001} & \textbf{0.195}\sd{0.001} \\
\xmark & \cmark & \cmark & 20.76\sd{0.10} & 0.719\sd{0.001} & 0.199\sd{0.001} \\
\cmidrule(r){1-3}\cmidrule(lr){4-6}
\cellcolor{ourgreen}\cmark & \cellcolor{ourgreen}\cmark & \cellcolor{ourgreen}\cmark & \cellcolor{ourgreen}\textbf{20.84}\sd{0.08} & \cellcolor{ourgreen}0.716\sd{0.001} & \cellcolor{ourgreen}\textbf{0.195}\sd{0.001} \\
\bottomrule
\end{tabular}
\end{table}

\rev{\rev{\noindent\textbf{Ablation study.}}
\rev{To examine the role of each component, \cref{tab:ablation} enables the three components alone and in combination.
Each component alone \rev{helps} in its own way: the orthogonal perturbation lowers the LPIPS by removing view-specific geometry, the conflict-aware correction gives the largest PSNR gain by protecting the reconstruction step, and color rescheduling raises the SSIM by suppressing the overfitting color.
The individual gains are small, but they combine, and all three together give the best PSNR and LPIPS.
Without color rescheduling, the two geometric components lower the PSNR slightly, which we attribute to the color overfitting of \cref{sec:mot_color}.}}

\label{sec:exp_cost}
\begin{table}[t]
\centering
\caption{\rev{\textbf{Wall-clock training time} (LLFF, 3 views, 10k iterations on a single GPU). $\times$3DGS: relative to 3DGS.}}
\label{tab:cost}\vspace{-0.5em}
\footnotesize
\setlength{\tabcolsep}{5pt}
\begin{tabular}{lcccccc}
\toprule
\multirow{2}{*}{Method} & \multicolumn{4}{c}{Training time (s)} & \multirow{2}{*}{$\times$3DGS} \\
\cmidrule(lr){2-5}
 & fern & horns & trex & mean & \\
\cmidrule(r){1-1}\cmidrule(lr){2-5}\cmidrule(lr){6-6}
3DGS~\cite{kerbl2023gaussian} & 270 & 265 & 262 & 266 & 1.00 \\
+SAP~\cite{seo2026sap} & 273 & 272 & 273 & 273 & 1.03 \\
+FASR~\cite{yun2026flat} & 279 & 275 & 291 & 282 & 1.06 \\
\cmidrule(r){1-1}\cmidrule(lr){2-5}\cmidrule(lr){6-6}
\rev{+\method{} w/o cache} & 552 & 555 & 548 & 552 & 2.08 \\
\rowcolor{ourgreen}
\textbf{\rev{+\method}~\rev{(Ours)}} & \textbf{288} & \textbf{296} & \textbf{294} & \textbf{293} & \textbf{1.10} \\
\bottomrule
\end{tabular}
\end{table}

\rev{\rev{\noindent\textbf{Training cost.}}
\rev{To verify that \method improves accuracy without a large increase in cost, \cref{tab:cost} measures the wall-clock training time.
\method takes $1.10\times$ the time of 3DGS, only 4\% more than FASR, so as a plug-in optimizer it delivers the gains of \cref{sec:exp_main} at little additional cost.
\rev{This efficiency comes from the gradient cache, which nearly halves the training time of the exact average gradient.}}}

\label{sec:exp_comp}
\begin{table}[!ht]
\centering
\caption{\rev{\textbf{Color rescheduling alone} on 3DGS (LLFF, 3 views). \rev{Color: color rescheduling.}}}
\label{tab:shr}\vspace{-0.5em}
\footnotesize
\setlength{\tabcolsep}{3pt}
\begin{tabular}{lccc}
\toprule
Method & PSNR$\uparrow$ & SSIM$\uparrow$ & LPIPS$\downarrow$ \\
\cmidrule(r){1-1}\cmidrule(lr){2-4}
3DGS~\cite{kerbl2023gaussian} & 19.74\sd{0.09} & 0.678\sd{0.001} & \textbf{0.214}\sd{0.001} \\
\cmidrule(r){1-1}\cmidrule(lr){2-4}
\rowcolor{ourgreen}\rev{+Color} & \textbf{19.90}\sd{0.13} & \textbf{0.683}\sd{0.001} & \textbf{0.214}\sd{0.001} \\
\bottomrule
\end{tabular}
\end{table}

\rev{\noindent\textbf{\rev{Effect of color rescheduling alone.}}}
\rev{The ablation above measures each component on top of FASR+SAP; to isolate color rescheduling from the perturbation, \cref{tab:shr} adds it alone to 3DGS. It still raises PSNR by $0.16$\,dB ($p<0.01$), so color rescheduling is effective on its own and generalizes beyond our perturbation.}

\section{\rev{Conclusion}}
\label{sec:conclusion}
\rev{In this work, we proposed \method, a plug-in optimizer that chooses the directions of flat-minima optimization from the gradients of all training views\rev{.}
\method perturbs each Gaussian along the component of its gradient on which the views disagree, removes the part of the flatness correction that opposes the clean gradient, and reschedules the overfitting color.
Across LLFF, Mip-NeRF\,360, \rev{Blender, and DTU}, \method consistently improves vanilla 3DGS and six sparse-view methods at nearly the cost of \rev{existing flat-minima optimizers}.
We believe that relating the perturbation to the other training views offers a starting point for flat-minima optimization wherever supervision is sparse.}

{
    \small
    \bibliographystyle{ieeenat_fullname}
    \bibliography{main}

@article{kerbl2023gaussian,
  title   = {{{3D} {Gaussian} Splatting for Real-Time Radiance Field Rendering}},
  author  = {Kerbl, Bernhard and Kopanas, Georgios and Leimk{\"u}hler, Thomas and Drettakis, George},
  journal = {ACM Transactions on Graphics},
  volume  = {42}, number = {4}, year = {2023}
}

@inproceedings{mildenhall2020nerf,
  title     = {{{NeRF}: Representing Scenes as Neural Radiance Fields for View Synthesis}},
  author    = {Mildenhall, Ben and Srinivasan, Pratul P. and Tancik, Matthew and Barron, Jonathan T. and Ramamoorthi, Ravi and Ng, Ren},
  booktitle = {ECCV}, year = {2020}
}

@article{mildenhall2019llff,
  title   = {{Local Light Field Fusion: Practical View Synthesis with Prescriptive Sampling Guidelines}},
  author  = {Mildenhall, Ben and Srinivasan, Pratul P. and Ortiz-Cayon, Rodrigo and Kalantari, Nima Khademi and Ramamoorthi, Ravi and Ng, Ren and Kar, Abhishek},
  journal = {ACM Transactions on Graphics}, volume = {38}, number = {4}, year = {2019}
}

@inproceedings{barron2022mipnerf360,
  title     = {{Mip-{NeRF} 360: Unbounded Anti-Aliased Neural Radiance Fields}},
  author    = {Barron, Jonathan T. and Mildenhall, Ben and Verbin, Dor and Srinivasan, Pratul P. and Hedman, Peter},
  booktitle = {CVPR}, year = {2022}
}

@inproceedings{zhang2018lpips,
  title     = {{The Unreasonable Effectiveness of Deep Features as a Perceptual Metric}},
  author    = {Zhang, Richard and Isola, Phillip and Efros, Alexei A. and Shechtman, Eli and Wang, Oliver},
  booktitle = {CVPR}, year = {2018}
}

@inproceedings{foret2021sam,
  title     = {{Sharpness-Aware Minimization for Efficiently Improving Generalization}},
  author    = {Foret, Pierre and Kleiner, Ariel and Mobahi, Hossein and Neyshabur, Behnam},
  booktitle = {ICLR}, year = {2021}
}

@inproceedings{zhuang2022gsam,
  title     = {{Surrogate Gap Minimization Improves Sharpness-Aware Training}},
  author    = {Zhuang, Juntang and Gong, Boqing and Yuan, Liangzhe and Cui, Yin and Adam, Hartwig and Dvornek, Nicha and Tatikonda, Sekhar and Duncan, James and Liu, Ting},
  booktitle = {ICLR}, year = {2022}
}

@inproceedings{yun2026flat,
  title     = {{Do Flat Minima Improve Sparse Novel View Synthesis?}},
  author    = {Yun, Youngsik and Gu, Dongjun and Uh, Youngjung},
  booktitle = {ECCV}, year = {2026}
}

@inproceedings{seo2026sap,
  title     = {{Improving Sparse-View {3DGS} Generalization via Flat Minima Optimization}},
  author    = {Seo, Kangmin and Hyun, Sangeek and Lee, MinKyu and Heo, Jae-Pil},
  booktitle = {ECCV}, year = {2026}
}

@inproceedings{park2025dropgaussian,
  title     = {{{DropGaussian}: Structural Regularization for Sparse-view {Gaussian} Splatting}},
  author    = {Park, Hyunwoo and Ryu, Gun and Kim, Wonjun},
  booktitle = {CVPR}, year = {2025}
}

@inproceedings{dropoutgs2025,
  title     = {{{DropoutGS}: Dropping Out {Gaussians} for Better Sparse-view Rendering}},
  author    = {Xu, Yexing and Wang, Longguang and Chen, Minglin and Ao, Sheng and Li, Li and Guo, Yulan},
  booktitle = {CVPR}, year = {2025}
}

@inproceedings{dropansh2026,
  title     = {{Dropping Anchor and Spherical Harmonics for Sparse-view {Gaussian} Splatting}},
  author    = {Fang, Shuangkang and Shen, I-Chao and Zhang, Xuanyang and Wang, Zesheng and Wang, Yufeng and Ding, Wenrui and Yu, Gang and Igarashi, Takeo},
  booktitle = {CVPR}, year = {2026}
}

@inproceedings{zhang2024corgs,
  title     = {{{CoR-GS}: Sparse-View {3D} {Gaussian} Splatting via Co-Regularization}},
  author    = {Zhang, Jiawei and Li, Jiahe and Yu, Xiaohan and Huang, Lei and Gu, Lin and Zheng, Jin and Bai, Xiao},
  booktitle = {ECCV}, year = {2024}
}

@inproceedings{segs2025,
  title     = {{Self-Ensembling {Gaussian} Splatting for Few-Shot Novel View Synthesis}},
  author    = {Zhao, Chen and Wang, Xuan and Zhang, Tong and Javed, Saqib and Salzmann, Mathieu},
  booktitle = {ICCV}, year = {2025}
}

@inproceedings{zhu2024fsgs,
  title     = {{{FSGS}: Real-Time Few-Shot View Synthesis Using {Gaussian} Splatting}},
  author    = {Zhu, Zehao and Fan, Zhiwen and Jiang, Yifan and Wang, Zhangyang},
  booktitle = {ECCV}, year = {2024}
}

@inproceedings{li2024dngaussian,
  title     = {{{DNGaussian}: Optimizing Sparse-View {3D} {Gaussian} Radiance Fields with Global-Local Depth Normalization}},
  author    = {Li, Jiahe and Zhang, Jiawei and Bai, Xiao and Zheng, Jin and Ning, Xin and Zhou, Jun and Gu, Lin},
  booktitle = {CVPR}, year = {2024}
}

@inproceedings{lu2024scaffold,
  title     = {{{Scaffold-GS}: Structured {3D} {Gaussians} for View-Adaptive Rendering}},
  author    = {Lu, Tao and Yu, Mulin and Xu, Linning and Xiangli, Yuanbo and Wang, Limin and Lin, Dahua and Dai, Bo},
  booktitle = {CVPR}, year = {2024}
}

@inproceedings{kim2026reward,
  title     = {{Reward Sharpness-Aware Fine-Tuning for Diffusion Models}},
  author    = {Kim, Kwanyoung and Sim, Byeongsu},
  booktitle = {CVPR}, year = {2026}
}

@inproceedings{niemeyer2022regnerf,
  title     = {{{RegNeRF}: Regularizing Neural Radiance Fields for View Synthesis from Sparse Inputs}},
  author    = {Niemeyer, Michael and Barron, Jonathan T. and Mildenhall, Ben and Sajjadi, Mehdi S. M. and Geiger, Andreas and Radwan, Noha},
  booktitle = {CVPR},
  year      = {2022}
}

@article{hochreiter1997flat,
  title   = {{Flat Minima}},
  author  = {Hochreiter, Sepp and Schmidhuber, J{\"u}rgen},
  journal = {Neural Computation},
  volume  = {9},
  number  = {1},
  pages   = {1--42},
  year    = {1997}
}

@inproceedings{keskar2017large,
  title     = {{On Large-Batch Training for Deep Learning: Generalization Gap and Sharp Minima}},
  author    = {Keskar, Nitish Shirish and Mudigere, Dheevatsa and Nocedal, Jorge and Smelyanskiy, Mikhail and Tang, Ping Tak Peter},
  booktitle = {ICLR},
  year      = {2017}
}

@article{li2024rwp,
  title   = {{Revisiting Random Weight Perturbation for Efficiently Improving Generalization}},
  author  = {Li, Tao and Tao, Qinghua and Yan, Weihao and Wu, Yingwen and Lei, Zehao and Fang, Kun and He, Mingzhen and Huang, Xiaolin},
  journal = {Transactions on Machine Learning Research},
  year    = {2024}
}

@inproceedings{kheradmand2024mcmc,
  title     = {{{3D} Gaussian Splatting as {Markov} Chain {Monte} {Carlo}}},
  author    = {Kheradmand, Shakiba and Rebain, Daniel and Sharma, Gopal and Sun, Weiwei and Tseng, Yang-Che and Isack, Hossam and Kar, Abhishek and Tagliasacchi, Andrea and Yi, Kwang Moo},
  booktitle = {NeurIPS},
  year      = {2024}
}

@inproceedings{liu2022looksam,
  title     = {{Towards Efficient and Scalable Sharpness-Aware Minimization}},
  author    = {Liu, Yong and Mai, Siqi and Chen, Xiangning and Hsieh, Cho-Jui and You, Yang},
  booktitle = {CVPR},
  pages     = {12360--12370},
  year      = {2022}
}

@inproceedings{zhou2026gdags,
  title     = {{Gradient-Direction-Aware Density Control for {3D} {Gaussian} Splatting}},
  author    = {Zhou, Zheng and Xiong, Yu-Jie and Zhang, Jia-Chen and Xia, Chun-Ming and Qiu, Xihe and Zhan, Hongjian},
  booktitle = {ICLR},
  year      = {2026}
}

@inproceedings{chen2022vit,
  title     = {{When Vision Transformers Outperform {ResNets} without Pre-training or Strong Data Augmentations}},
  author    = {Chen, Xiangning and Hsieh, Cho-Jui and Gong, Boqing},
  booktitle = {ICLR},
  year      = {2022}
}

@inproceedings{cha2021swad,
  title     = {{{SWAD}: Domain Generalization by Seeking Flat Minima}},
  author    = {Cha, Junbum and Chun, Sanghyuk and Lee, Kyungjae and Cho, Han-Cheol and Park, Seunghyun and Lee, Yunsung and Park, Sungrae},
  booktitle = {NeurIPS},
  year      = {2021}
}

@inproceedings{bahri2022sam,
  title     = {{Sharpness-Aware Minimization Improves Language Model Generalization}},
  author    = {Bahri, Dara and Mobahi, Hossein and Tay, Yi},
  booktitle = {ACL},
  pages     = {7360--7371},
  year      = {2022}
}

@article{wang2004ssim,
  title   = {{Image Quality Assessment: From Error Visibility to Structural Similarity}},
  author  = {Wang, Zhou and Bovik, Alan C. and Sheikh, Hamid R. and Simoncelli, Eero P.},
  journal = {IEEE Transactions on Image Processing},
  volume  = {13},
  number  = {4},
  pages   = {600--612},
  year    = {2004}
}

@inproceedings{kingma2015adam,
  title     = {{Adam: A Method for Stochastic Optimization}},
  author    = {Kingma, Diederik P. and Ba, Jimmy},
  booktitle = {ICLR},
  year      = {2015}
}

@article{wilcoxon1945,
  title   = {{Individual Comparisons by Ranking Methods}},
  author  = {Wilcoxon, Frank},
  journal = {Biometrics Bulletin},
  volume  = {1},
  number  = {6},
  pages   = {80--83},
  year    = {1945}
}

@inproceedings{jensen2014dtu,
  author    = {Jensen, Rasmus and Dahl, Anders and Vogiatzis, George and Tola, Engin and Aan{\ae}s, Henrik},
  title     = {{Large Scale Multi-view Stereopsis Evaluation}},
  booktitle = {CVPR},
  pages     = {406--413},
  year      = {2014}
}
}

\clearpage

\clearpage
\maketitlesupplementary
\appendix

\startcontents[appendix]
\section*{Appendix Contents}
\printcontents[appendix]{}{1}{\setcounter{tocdepth}{2}}

\section{Extended Related Work}
\label{app:related}
\subsection{Sparse-View 3DGS}
The methods we build on fall into two groups.
The first adds supervision: \rev{DNGaussian~\cite{li2024dngaussian} regularizes the rendered depth toward a monocular prior with hard and soft depth losses under a global-local normalization; FSGS~\cite{zhu2024fsgs} densifies Gaussians during training by proximity-guided unpooling, supervises depth from a monocular estimator, and synthesizes pseudo views.}
The second constrains the optimization without external priors: CoR-GS~\cite{zhang2024corgs} trains two models in parallel and co-prunes Gaussians on which they disagree; \rev{SE-GS~\cite{segs2025} perturbs one model by rendering uncertainty and regularizes a second, unperturbed model toward the perturbed copies at pseudo views;} DropGaussian~\cite{park2025dropgaussian} randomly drops Gaussians during training so that the remaining ones receive larger gradients, \rev{DropoutGS~\cite{dropoutgs2025} randomly deactivates Gaussians during training and supervises the reduced rendering with that of the full model,} and DropAnSH-GS~\cite{dropansh2026} additionally drops high-degree SH coefficients.
All of these change what is rendered or what is supervised; none changes how the gradient is used.
\method is complementary to both groups, and \cref{tab:plugin_llff} shows that it adds to each of CoR-GS, SE-GS, DropGaussian, DNGaussian, and FSGS.
Scaffold-GS~\cite{lu2024scaffold} organizes Gaussians around anchors and has no SH color; \VAS and \CAU without \SHR improve it by the largest margin of all code bases (\cref{tab:plugin_llff}).

\rev{\subsection{Related Gradient Decompositions}
Gradient decompositions also appear in other SAM variants and in 3DGS.
GSAM~\cite{zhuang2022gsam} decomposes the gradient of the clean loss against the gradient of the perturbed loss and ascends along the orthogonal component, which lowers the surrogate gap without changing the descent on the perturbed loss.
LookSAM~\cite{liu2022looksam} splits the SAM gradient into a component parallel to the SGD gradient and an orthogonal component, and reuses the latter across iterations to reduce the cost of SAM.
In 3DGS, GDAGS~\cite{zhou2026gdags} measures the directional coherence of the gradients that each Gaussian receives and uses it to control densification in dense-view reconstruction.
None of them relates the perturbation of a flat-minima optimizer to the gradients of the other training views.}

\rev{\section{Proof of Proposition 1}
\label{app:proofs}
\noindent\textbf{Proposition 1.}
\emph{For every Gaussian $i$, $\langle\uv_i,\gv_i\rangle=\|\gv_i\|^2+\lambda\max(\langle\dv_i,\gv_i\rangle,0)\ge\|\gv_i\|^2$, and the component of $\uv_i$ orthogonal to $\gv_i$ equals that of \rev{$\gv_i+\lambda\dv_i$}, $\lambda(\dv_i-\dv_i^{\gv})$.}

\begin{proof}
Let $\dv_i^{\gv}=\langle\dv_i,\gv_i\rangle\gv_i/\|\gv_i\|^2$ be the orthogonal projection of $\dv_i$ onto $\gv_i$, so that
\begin{equation*}
  \langle\dv_i^{\gv},\gv_i\rangle=\langle\dv_i,\gv_i\rangle,\qquad\langle\dv_i-\dv_i^{\gv},\gv_i\rangle=0.
\end{equation*}
We consider the two cases of \cref{eq:cau}.
If $\langle\dv_i,\gv_i\rangle\ge0$, then $\uv_i=\gv_i+\lambda\dv_i$, and
\begin{equation*}
  \langle\uv_i,\gv_i\rangle=\|\gv_i\|^2+\lambda\langle\dv_i,\gv_i\rangle.
\end{equation*}
Otherwise, $\uv_i=\gv_i+\lambda(\dv_i-\dv_i^{\gv})$, and
\begin{equation*}
  \langle\uv_i,\gv_i\rangle=\|\gv_i\|^2+\lambda\langle\dv_i-\dv_i^{\gv},\gv_i\rangle=\|\gv_i\|^2.
\end{equation*}
Since $\lambda\ge0$, the two cases together give
\begin{equation*}
  \langle\uv_i,\gv_i\rangle=\|\gv_i\|^2+\lambda\max(\langle\dv_i,\gv_i\rangle,0)\ge\|\gv_i\|^2.
\end{equation*}
For the second claim, we write the two cases as
\begin{equation*}
  \uv_i=\begin{cases}\bigl(\gv_i+\lambda\dv_i^{\gv}\bigr)+\lambda(\dv_i-\dv_i^{\gv}), & \langle\dv_i,\gv_i\rangle\ge0,\\ \gv_i+\lambda(\dv_i-\dv_i^{\gv}), & \text{otherwise}.\end{cases}
\end{equation*}
In both cases, the first term is parallel to $\gv_i$ and the second is orthogonal to it.
The orthogonal component of $\uv_i$ is therefore $\lambda(\dv_i-\dv_i^{\gv})$, as for \rev{$\gv_i+\lambda\dv_i$} in \cref{eq:rescale}, which proves \cref{prop:cau}.
\end{proof}

\paragraph{Remark.}
\Cref{prop:cau} concerns the direction $\uv_i$ passed to the optimizer and holds exactly for a gradient-descent step along it.
Adam rescales each coordinate of $\uv_i$ by its second-moment estimate, so the actual Adam step need not satisfy the same bound.}

\section{Scope and Limitations}
\label{app:scope}

\rev{\subsection{Limitations}
The gain of \method shrinks as the number of views grows, and with 9 LLFF views color rescheduling should be switched off.
With 12 Mip-NeRF\,360 views, the cached average becomes stale for the Gaussians with the largest gradients (see below).}

\subsection{Color Rescheduling and View Count}
Adding \SHR on top of \VAS and \CAU helps with 3 LLFF views and 12 Mip-NeRF\,360 views, is neutral with 6 LLFF, 24 Mip-NeRF\,360, and 8 Blender views, and no longer helps with 9 LLFF views, where \VAS and \CAU alone is the best configuration (\cref{tab:ablation69}).
This matches its motivation (\cref{sec:motivation}): the view-dependent color overfits when it is fitted from \rev{few} directions, and with enough directions freezing it early only removes capacity.
Color rescheduling should therefore be read as a component for the few-view regime; all main-paper results use the same schedule for every view count.

\subsection{Accuracy of the Cached \rev{Average} Gradient}
\label{app:cache}
We compared the cached \rev{average} gradient of \cref{sec:method} with the exact one during training, recomputing the exact \rev{average} from all training views every 50 iterations after the warm-up.
\rev{On LLFF, the} cached entries are on average 1.8 (3 views) and 3.5 (6 views) iterations old, and the per-Gaussian median cosine between cached and exact \rev{average} is 0.99; the Gaussians with the largest gradients, which dominate the mean, agree almost perfectly (median cosine 1.00).
\rev{On Mip-NeRF\,360 with 12 views, the} entries are 7 iterations old and the typical Gaussian still agrees (median 0.91), but the 1\% of Gaussians with the largest gradients, which carry 82\% of the energy of the mean, do not (median 0.17): these are opaque, well-observed Gaussians of the central object whose per-view position gradient changes sign between visits, not floaters.
We use the cache everywhere.

\section{Experimental Details}
\label{app:details}
\subsection{Protocol}
\rev{All experiments use the training protocol of FASR~\cite{yun2026flat} and the seed fixed by each code base, with 5 identical repeats per configuration, and we report the mean and standard deviation over repeats of the scene-averaged metric.}
Paired differences are computed per (scene, repeat) and tested with a two-sided Wilcoxon signed-rank test.
\subsection{Datasets and Views}
LLFF: 8 scenes, 1/8 resolution, 10k iterations, training views chosen as in FASR and the test views every eighth image; Mip-NeRF\,360: 7 scenes, 1/8 resolution, 10k iterations, 12 (and 24) training views sampled uniformly over the capture, the 12-view multi-view-stereo point cloud as initialization for both view counts; Blender: 8 scenes, $400\times400$, white background, random 10k-point initialization, 10k iterations, 8 training views.

\subsection{Hyperparameters}
Sharpness-aware optimization starts at iteration 3k and perturbs positions only, with $\gamma=0.95$ and the per-Gaussian radius factor $\tau_i$ and correction weight $\bar\gamma_i$ of FASR, and $\sigma=0.1$ \rev{for SAP}; \method uses $\rho=2$ (FASR: $\rho=1$) \rev{and adds the SAP noise with the same $\sigma$, without the periodic re-initialization of SAP,} the \rev{average} gradient over all training views including the current one, and the SH schedule $(T_{\mathrm f},T_{\mathrm r},T_{\mathrm e})=(2\text{k},4\text{k},6\text{k})$ with a 2k-iteration re-fitting window after reinitialization.
\subsection{Plug-in Implementation}
\rev{In every code base, the} optimizer hooks into the training loop at the same two places: after the clean backward pass (where the cached \rev{average} gradient is read, the current gradient is cached, and the perturbation is applied) and after the perturbed backward pass (where \CAU forms the update and the parameters are restored).
CoR-GS applies the optimizer to both of its co-regularized models, SE-GS to the primary model, DropGaussian reuses the gradient of its dropped render as the clean gradient, FSGS keeps its pseudo-view and depth losses unchanged, and DNGaussian, whose official schedule trains for 6k iterations, uses \VAS and \CAU without \SHR.
Scaffold-GS perturbs its anchor positions, applies \CAU to the offsets, uses the alpha-normalized expected depth of its renderer for the geometry-aware radius, and, having no SH color, uses \VAS and \CAU without \SHR; its \rev{average} gradient is computed exactly rather than cached.
For Mip-NeRF\,360, each code uses its $360^\circ$ scripts and settings (co-pruning threshold 20 for CoR-GS and SE-GS, noise $0.02$/$0.001$ for SE-GS, and pseudo-depth weight $0.03$ for FSGS).

\section{Additional Quantitative Results}
\label{app:results}
\rev{\subsection{Protocol on Blender and DTU}
For \cref{tab:plugin_blender_dtu}, we add the optimizers to 3DGS and to the official Blender and DTU recipes of CoR-GS and DNGaussian.
DTU uses 3 input views on the 15 scans of RegNeRF~\cite{niemeyer2022regnerf}, evaluated with the object masks as in the baselines.
The DTU +SAP row of CoR-GS is averaged over 14 of the 15 scans, since scan110 has no 3-view MVS initialization in the official preprocessed release.
DNGaussian follows its official 6k-iteration Blender recipe; the others use 10k.}

\rev{\subsection{Magnitude of the Orthogonal Component}
\Cref{tab:e2} reports the mean of $\|\mathbf{g}_{\perp}\|/\|\gv\|$ of \cref{sec:mot_pert} over the Gaussians with a nonzero gradient, measured with plain 3DGS and the gradient cache over iterations 3k--10k on the 8 LLFF scenes. The orthogonal component carries about half of the gradient for every number of views (0.47, 0.55, and 0.57 with 3, 6, and 9 views), with almost no variation across runs, so it does not vanish as views are added.}
\begin{table}[!ht]
\centering\footnotesize
\caption{\rev{\textbf{Size of the orthogonal component} versus the number of input views (LLFF, 8 scenes). Mean of $\|\mathbf{g}_{\perp}\|/\|\gv\|$ over the Gaussians with a nonzero gradient, for three runs of plain 3DGS with the gradient cache.}}
\label{tab:e2}
\setlength{\tabcolsep}{9pt}
\begin{tabular}{ccc}
\toprule
Views & Run & $\|\mathbf{g}_{\perp}\|/\|\gv\|$ \\
\cmidrule(r){1-1}\cmidrule(lr){2-2}\cmidrule(lr){3-3}
\multirow{3}{*}{3} & 1 & 0.468 \\
 & 2 & 0.467 \\
 & 3 & 0.468 \\
\cmidrule(r){1-1}\cmidrule(lr){2-2}\cmidrule(lr){3-3}
\multirow{3}{*}{6} & 1 & 0.546 \\
 & 2 & 0.545 \\
 & 3 & 0.545 \\
\cmidrule(r){1-1}\cmidrule(lr){2-2}\cmidrule(lr){3-3}
\multirow{3}{*}{9} & 1 & 0.569 \\
 & 2 & 0.569 \\
 & 3 & 0.569 \\
\bottomrule
\end{tabular}
\end{table}
\rev{\subsection{Effect of the Direction without Noise}
\Cref{tab:direction_nosap} repeats the experiment of \cref{tab:direction} with the SAP noise off ($\sigma=0$). The ordering of the two directions is unchanged: the parallel direction collapses at $\rho=4$, below plain 3DGS, whereas the orthogonal direction stays at the level of FASR at every radius. Compared with \cref{tab:direction}, the noise raises the PSNR of the orthogonal perturbation at every radius, most at $\rho=1$ (20.45 to 20.66\,dB), which is why \method keeps it (\cref{sec:mot_pert}).}
\begin{table}[!ht]
\centering
\caption{\rev{\textbf{Direction of the perturbation without the SAP noise} (LLFF, 3 views; mean and standard deviation over 5 runs). The FASR perturbation is redirected along the parallel or orthogonal component with $\sigma=0$, to be compared with FASR; \cref{tab:direction} is the same experiment with the noise on. Bold: best per column.}}
\label{tab:direction_nosap}
\footnotesize
\setlength{\tabcolsep}{3pt}
\begin{tabular}{lcccc}
\toprule
Direction & $\rho$ & PSNR$\uparrow$ & SSIM$\uparrow$ & LPIPS$\downarrow$ \\
\cmidrule(r){1-1}\cmidrule(lr){2-2}\cmidrule(lr){3-5}
3DGS~\cite{kerbl2023gaussian} & -- & 19.74\sd{0.09} & 0.678\sd{0.001} & 0.214\sd{0.001} \\
+FASR~\cite{yun2026flat} & 1 & 20.67\sd{0.06} & 0.715\sd{0.001} & 0.198\sd{0.001} \\
\cmidrule(r){1-1}\cmidrule(lr){2-2}\cmidrule(lr){3-5}
\multirow{3}{*}{Parallel ($\gpar$)} & 1 & 20.62\sd{0.04} & \textbf{0.717}\sd{0.001} & 0.197\sd{0.001} \\
 & 2 & 20.66\sd{0.02} & 0.716\sd{0.002} & 0.199\sd{0.001} \\
 & 4 & 19.41\sd{0.08} & 0.612\sd{0.001} & 0.287\sd{0.001} \\
\cmidrule(r){1-1}\cmidrule(lr){2-2}\cmidrule(lr){3-5}
\multirow{3}{*}{Orthogonal ($\gperp$)} & 1 & 20.45\sd{0.10} & 0.704\sd{0.001} & 0.195\sd{0.001} \\
 & 2 & \textbf{20.70}\sd{0.03} & 0.715\sd{0.001} & \textbf{0.194}\sd{0.001} \\
 & 4 & 20.67\sd{0.05} & 0.709\sd{0.001} & \textbf{0.194}\sd{0.001} \\
\bottomrule
\end{tabular}
\end{table}

\subsection{Ablation with 6 and 9 Views}
\begin{table}[!ht]
\centering
\caption{\rev{\textbf{Ablation of the three components} (LLFF, 6 and 9 views). Pert.: orthogonal perturbation; Corr.: conflict-aware correction; Color: color rescheduling. Bold: best per block.}}
\label{tab:ablation69}
\footnotesize
\setlength{\tabcolsep}{3.5pt}
\begin{tabular}{lccc ccc}
\toprule
Views & Pert. & Corr. & Color & PSNR$\uparrow$ & SSIM$\uparrow$ & LPIPS$\downarrow$ \\
\cmidrule(r){1-1}\cmidrule(lr){2-4}\cmidrule(lr){5-7}
\multirow{8}{*}{6}
 & \xmark & \xmark & \xmark & 24.08\sd{0.07} & 0.811\sd{0.001} & 0.143\sd{0.001} \\
 & \cmark & \xmark & \xmark & 24.07\sd{0.06} & 0.814\sd{0.001} & 0.137\sd{0.001} \\
 & \xmark & \cmark & \xmark & 24.08\sd{0.05} & 0.811\sd{0.001} & 0.142\sd{0.001} \\
 & \xmark & \xmark & \cmark & 24.07\sd{0.04} & 0.811\sd{0.001} & 0.149\sd{0.001} \\
 & \cmark & \cmark & \xmark & 24.11\sd{0.07} & 0.814\sd{0.001} & \textbf{0.135}\sd{0.001} \\
 & \cmark & \xmark & \cmark & \textbf{24.14}\sd{0.06} & \textbf{0.815}\sd{0.002} & 0.140\sd{0.001} \\
 & \xmark & \cmark & \cmark & 24.04\sd{0.10} & 0.811\sd{0.002} & 0.148\sd{0.001} \\
\cmidrule(lr){2-4}\cmidrule(lr){5-7}
\rowcolor{ourgreen}  & \cmark & \cmark & \cmark & 24.11\sd{0.06} & \textbf{0.815}\sd{0.001} & 0.140\sd{0.001} \\
\midrule
\multirow{8}{*}{9}
 & \xmark & \xmark & \xmark & 25.21\sd{0.04} & 0.844\sd{0.000} & 0.127\sd{0.001} \\
 & \cmark & \xmark & \xmark & 25.31\sd{0.05} & 0.848\sd{0.001} & 0.119\sd{0.001} \\
 & \xmark & \cmark & \xmark & 25.25\sd{0.03} & 0.845\sd{0.001} & 0.126\sd{0.001} \\
 & \xmark & \xmark & \cmark & 25.13\sd{0.06} & 0.842\sd{0.001} & 0.134\sd{0.001} \\
 & \cmark & \cmark & \xmark & \textbf{25.35}\sd{0.04} & \textbf{0.849}\sd{0.001} & \textbf{0.118}\sd{0.001} \\
 & \cmark & \xmark & \cmark & 25.21\sd{0.05} & 0.847\sd{0.001} & 0.125\sd{0.001} \\
 & \xmark & \cmark & \cmark & 25.14\sd{0.02} & 0.843\sd{0.001} & 0.132\sd{0.001} \\
\cmidrule(lr){2-4}\cmidrule(lr){5-7}
\rowcolor{ourgreen}  & \cmark & \cmark & \cmark & 25.21\sd{0.05} & 0.847\sd{0.001} & 0.124\sd{0.001} \\
\bottomrule
\end{tabular}
\end{table}

\Cref{tab:ablation69} repeats the ablation of \cref{sec:exp_ablation} with 6 and 9 LLFF views: the gain of \SHR shrinks as the growing number of views carries more information about the view-dependent color, and with 9 views \VAS and \CAU alone is best (\cref{app:scope}).

\section{Additional Qualitative Results}
\label{app:qual}
\subsection{Qualitative Ablation}
\begin{figure*}[t]
  \centering
  \includegraphics[width=0.9\linewidth]{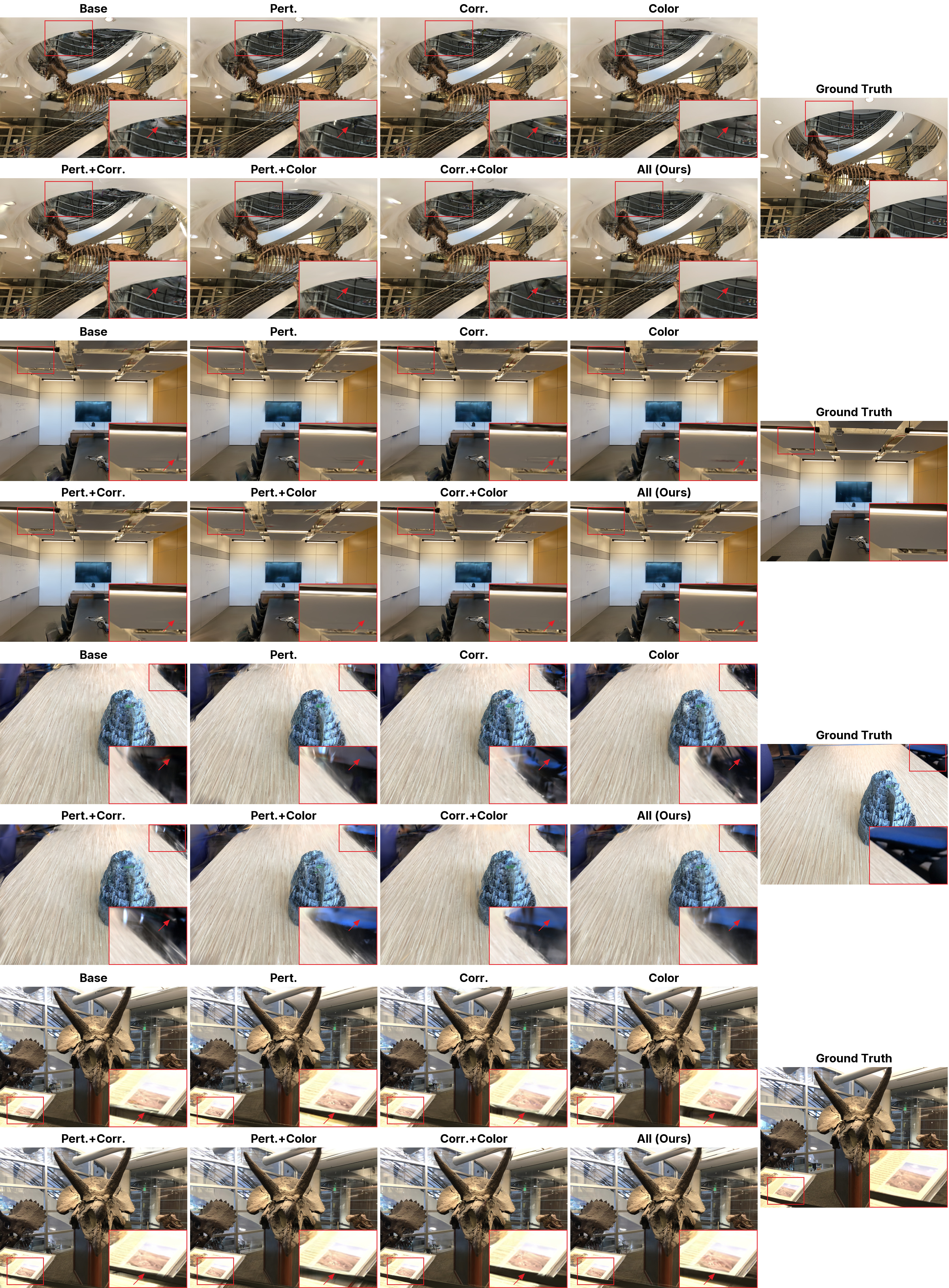}
  \caption{\rev{\textbf{Qualitative ablation of the three components} (LLFF, 3 views), the visual counterpart of \cref{tab:ablation}. Pert.: orthogonal perturbation; Corr.: conflict-aware correction; Color: color rescheduling.}}
  \label{fig:qual_abl}
\end{figure*}

\rev{\Cref{fig:qual_abl} shows the same configurations as \cref{tab:ablation} on novel views of four LLFF scenes.}
\subsection{Qualitative Plug-in Results}
\begin{figure*}[t]
  \centering
  \includegraphics[width=\linewidth]{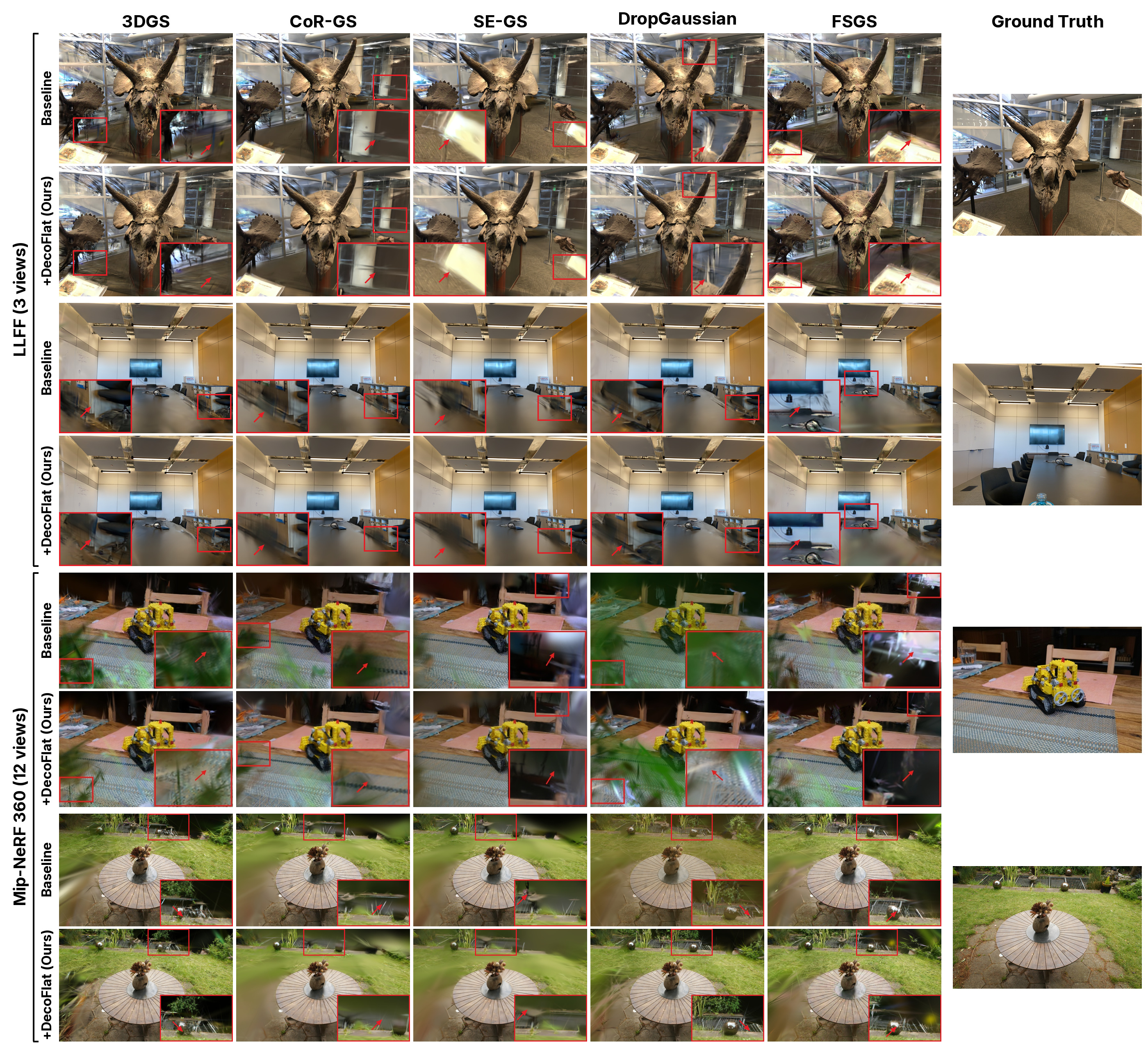}
  \caption{\textbf{Qualitative plug-in results.} Each method is trained with its official code (Baseline) and with \method attached. The floaters and blurred regions of each baseline are reduced when \method is attached, on both LLFF and Mip-NeRF 360.}
  \label{fig:qual_plugin}
\end{figure*}

\begin{figure*}[t]
  \centering
  \includegraphics[width=0.95\linewidth]{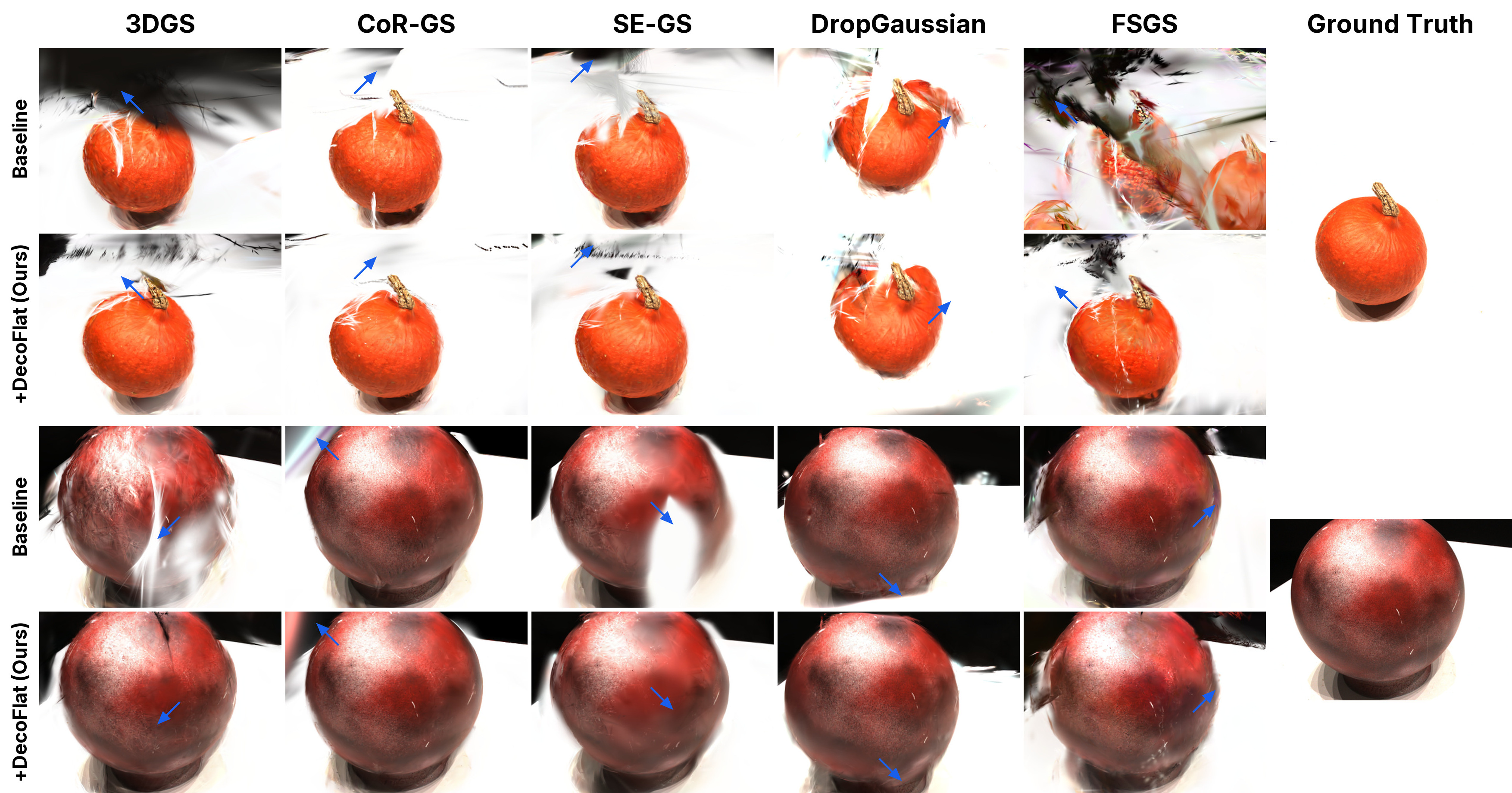}
  \caption{\textbf{Qualitative plug-in results on DTU} (3 views), same layout as \cref{fig:qual_plugin}. Arrows mark the artifact that is reduced. The renderings are shown without the object mask.}
  \label{fig:qual_dtu}
\end{figure*}

\begin{figure*}[t]
  \centering
  \includegraphics[width=0.95\linewidth]{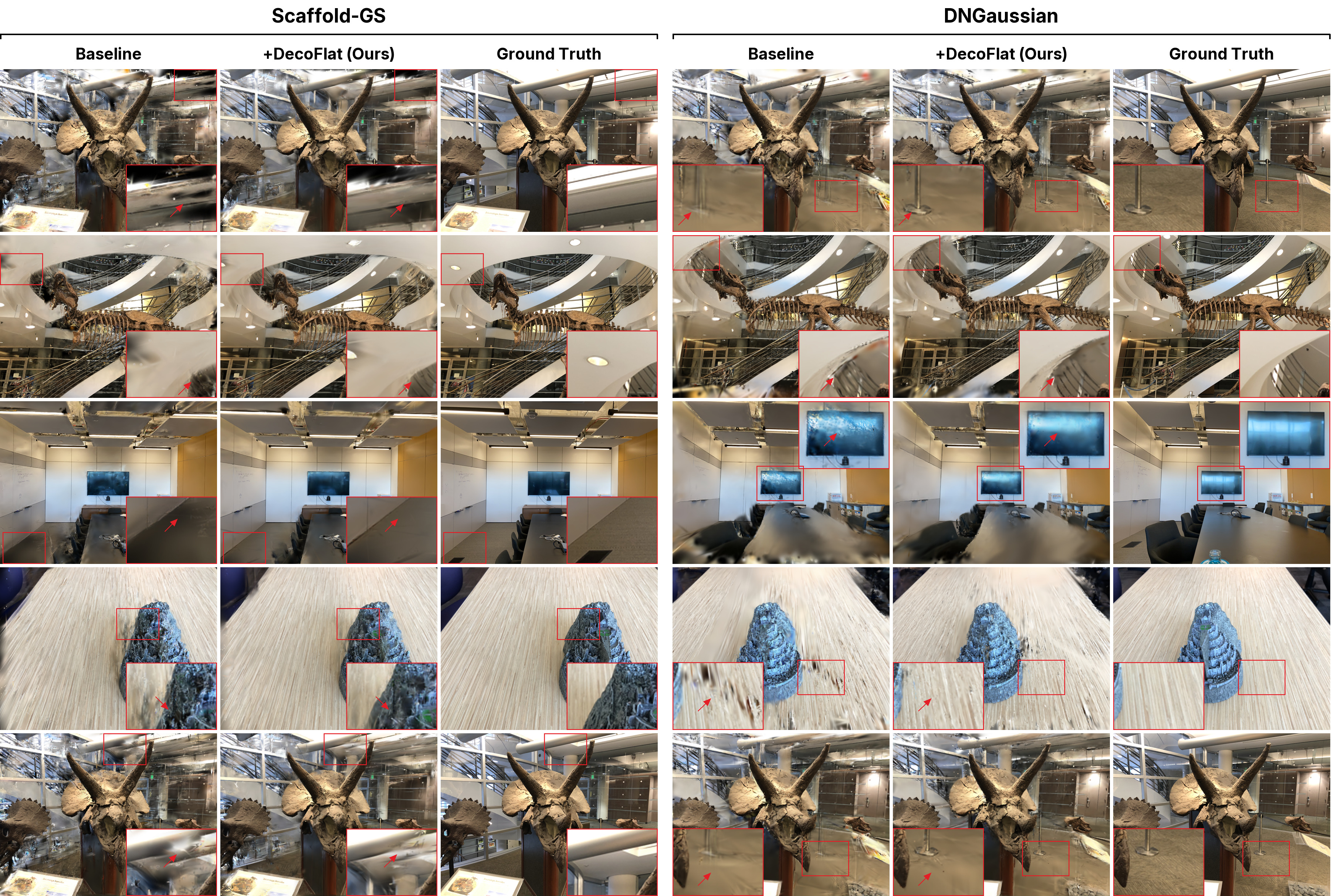}
  \caption{\textbf{Qualitative plug-in results for Scaffold-GS and DNGaussian} on LLFF (3 views). For each code base, a row shows the baseline, the same code with \method attached, and the ground truth; the test view is chosen per code base.}
  \label{fig:qual_plugin_app}
\end{figure*}

\Cref{fig:qual_plugin,fig:qual_plugin_app} show novel views of the plug-in experiment of \cref{tab:plugin_llff}: the view-specific artifacts of each baseline disappear when \method is attached to the same code.
\Cref{fig:qual_dtu} shows the same comparison on two DTU scans: the floaters that cover the object or the background in the baselines are reduced with \method.
\subsection{Comparison with Other Optimizers}
\begin{figure*}[t]
  \centering
  \includegraphics[width=\linewidth]{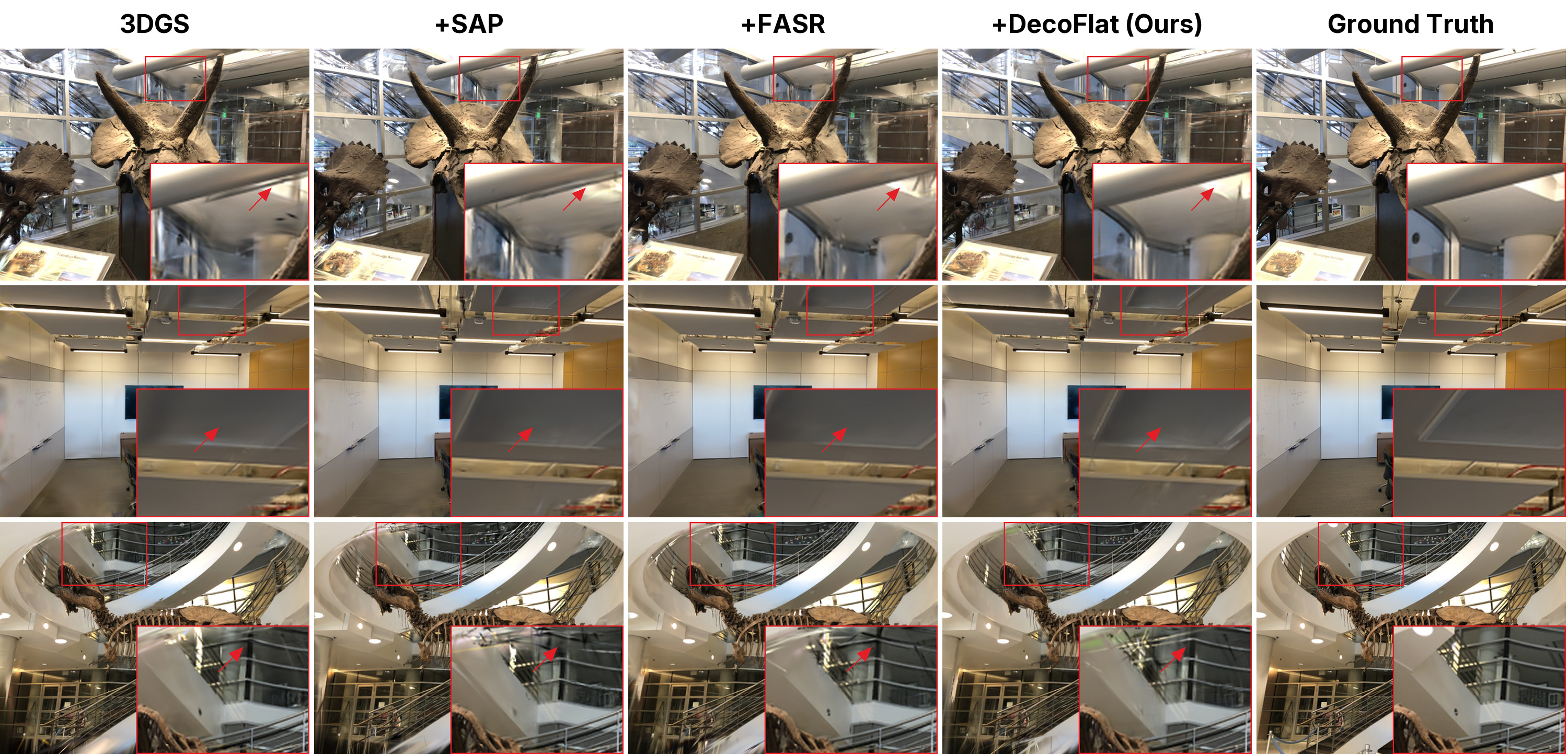}
  \caption{\textbf{Additional qualitative comparison on LLFF} (3 views) on the 3DGS backbone, with the columns of \cref{fig:qual}.}
  \label{fig:qual_llff_app}
\end{figure*}

\begin{figure*}[t]
  \centering
  \includegraphics[width=\linewidth]{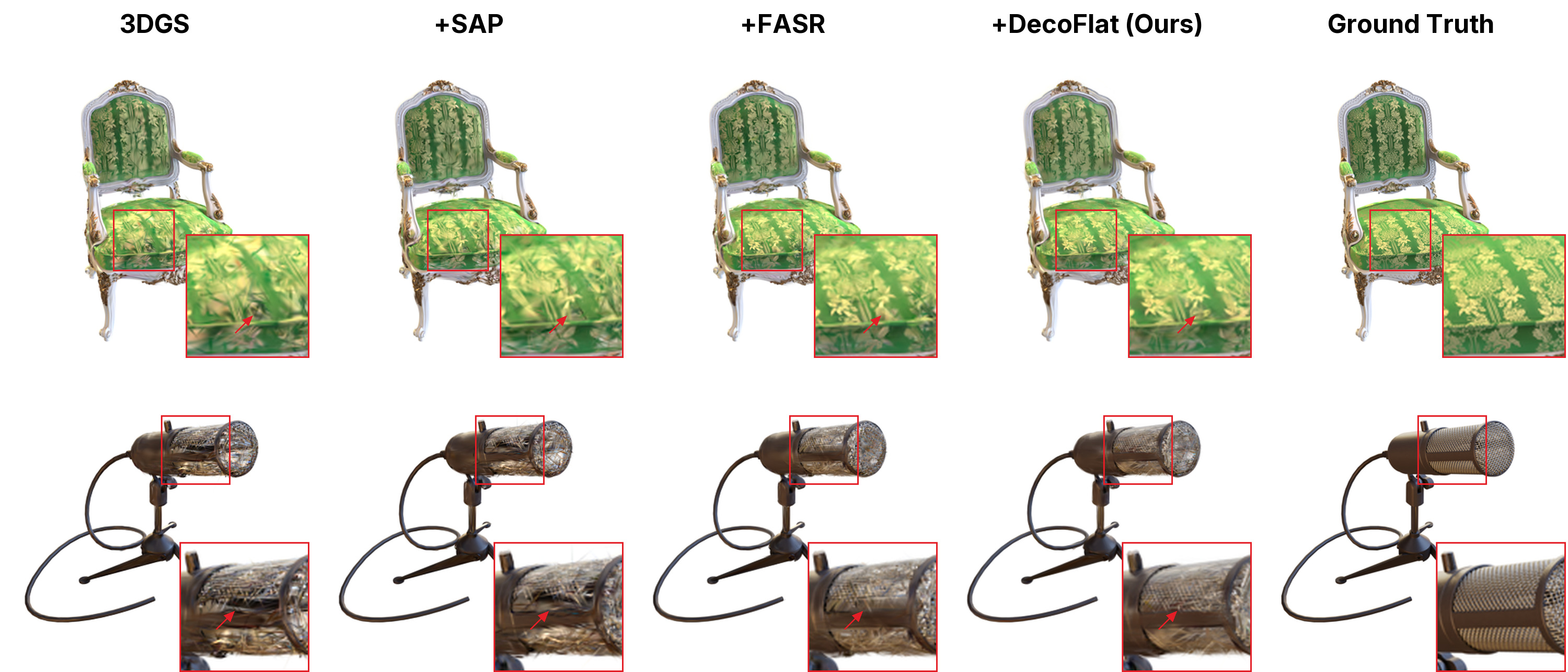}
  \caption{\textbf{Additional qualitative comparison on Blender} (8 views) on the 3DGS backbone, with the columns of \cref{fig:qual}.}
  \label{fig:qual_blender_app}
\end{figure*}

\begin{figure*}[t]
  \centering
  \includegraphics[width=\linewidth]{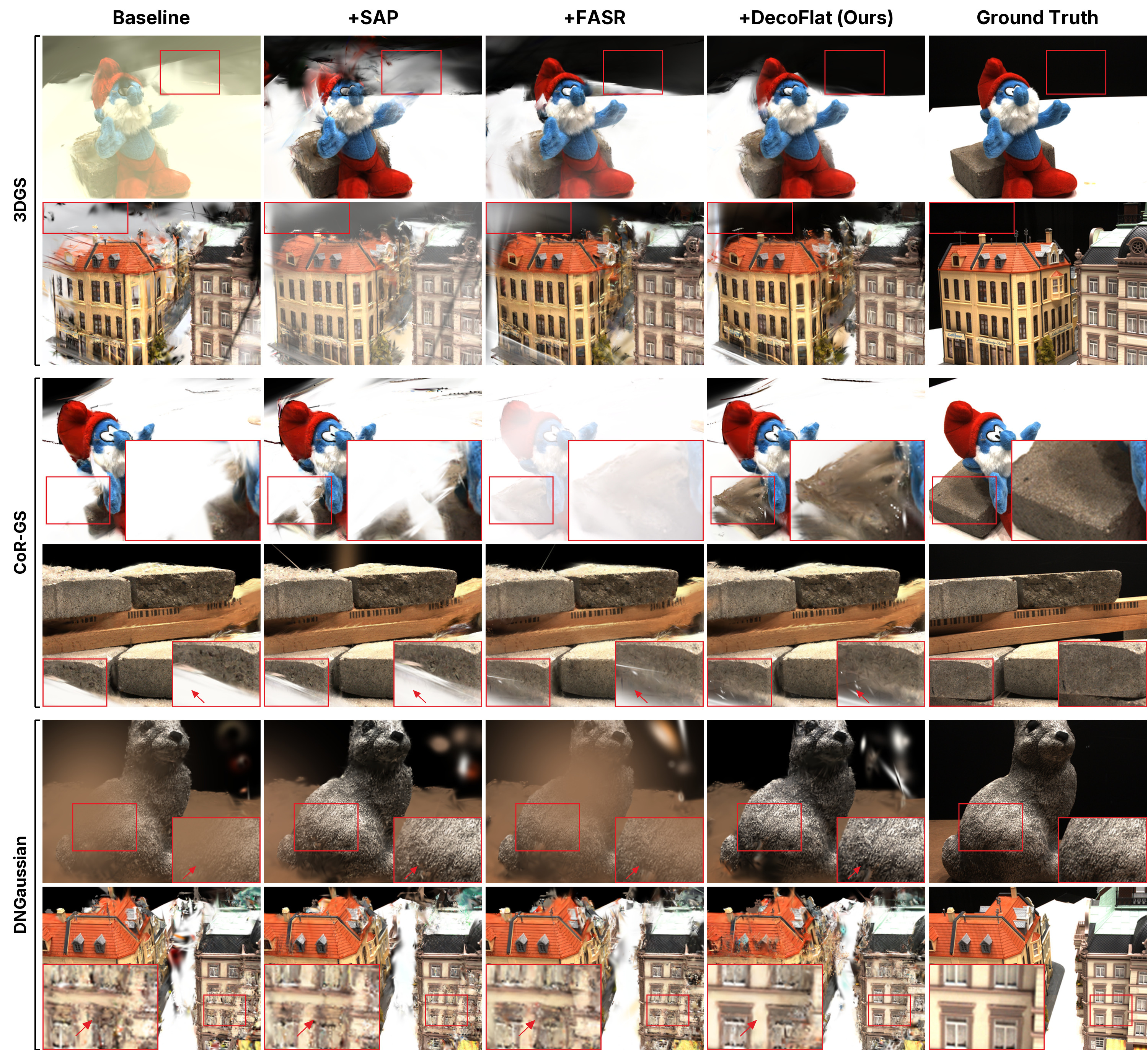}
  \caption{\textbf{Qualitative comparison of the optimizers on DTU} (3 views) for the three code bases of \cref{tab:plugin_blender_dtu}: each baseline and the same code with SAP, FASR, and \method attached. The renderings are shown without the object mask.}
  \label{fig:qual_dtu_methods}
\end{figure*}

\Cref{fig:qual_llff_app,fig:qual_blender_app,fig:qual_dtu_methods} compare \method with the other flat-minima optimizers on the same code base.
On LLFF, the floaters and ghosting of 3DGS that remain in part with SAP or FASR are reduced with \method (\cref{fig:qual_llff_app}).
On Blender and DTU, the boxed regions show floaters and haze of the baseline that are reduced when \method is attached (\cref{fig:qual_blender_app,fig:qual_dtu_methods}).

\end{document}